\documentclass[twocolumn]{article}

\usepackage{amsmath,amssymb,amsfonts}
\usepackage{bbold}
\usepackage{algorithmic}
\usepackage{graphicx}
\usepackage{textcomp}
\usepackage{xcolor}
\usepackage{subcaption}
\usepackage{booktabs,multirow,multicol} 
\usepackage{rotating} 
\usepackage[inline]{enumitem}
\usepackage{scalefnt}
\usepackage{tikz}
\usepackage{url}
\usepackage{hyperref}   
\usepackage{authblk}
\usepackage{orcidlink}
\usepackage{natbib}

\definecolor{sns-blue}{rgb}{0.2980392156862745, 0.4470588235294118, 0.6901960784313725}
\definecolor{sns-orange}{rgb}{0.867, 0.518, 0.322}
\definecolor{sns-green}{rgb}{0.333, 0.659, 0.408}
\definecolor{sns-red}{rgb}{0.769, 0.306, 0.322}
\definecolor{sns-purple}{rgb}{0.506, 0.447, 0.702}

\definecolor{brca-normal}{rgb}{0.2980392156862745, 0.4470588235294118, 0.6901960784313725}
\definecolor{brca-luma}{rgb}{0.867, 0.518, 0.322}
\definecolor{brca-lumb}{rgb}{0.333, 0.659, 0.408}
\definecolor{brca-basal}{rgb}{0.769, 0.306, 0.322}
\definecolor{brca-her2}{rgb}{0.506, 0.447, 0.702}

\newcommand{\sign}{\text{sgn}}
\newcommand{\corr}{\text{corr}}

\begin{document}

\title{Biologically Informed Deep Neural Networks for Multi-Omic Integration, Pathway Activity Inference and Risk Stratification in Cancer}

\author[kcl,astar,uom]{Pedro Henrique da Costa Avelar\orcidlink{0000-0002-0347-7002}}

\author[szu]{Le Ou-Yang\,\orcidlink{0000-0003-4007-4568}}

\author[astar]{Min Wu\orcidlink{0000-0003-0977-3600}}

\author[kcl]{Sophia Tsoka\orcidlink{0000-0001-8403-1282}}

\affil[kcl]{Department of Informatics, King's College London, London WC2B 4BG, United Kingdom}
\affil[astar]{Institute for Infocomm Research, Agency for Science, Technology and Research (A*STAR), Singapore 138632, Singapore}
\affil[szu]{College of Electronics and Information Engineering, Shenzhen University, Shenzhen 518054, Guangdong, China}
\affil[uom]{Current Affiliation: Division of Informatics, Imaging and Data Science, Faculty of Biology, Medicine and Health, University of Manchester}

\date{2025}

\maketitle

\begin{abstract}
Integrating complex, multi-omics data presents significant challenges. Existing approaches often face a trade-off between model interpretability and representational capacity, with most either relying on post-hoc interpretation or use linear models that may overlook complex interactions.
We report Pathway Activity Autoencoders for the multi-omics setting, which embed prior knowledge via pathway-informed architectural constraints, fostering interpretability, while preserving representational power.
Our multi-omic framework is applied in the context of breast cancer and is evaluated in survival prediction and subtype classification with results indicating a positive effect of integration. We conduct analysis of individual omics layer impact on end-task performance, revealing that gene, protein, and microRNA expression layers provide the strongest contribution. Repeatability studies indicate that, while dropout improves model robustness and consistency, excessive regularisation can reduce predictive performance. Finally, visualizations of the learned feature space illustrate the framework's intrinsic transparency and clinical relevance.
The results underscore the value of multi-omic integration and delineate the impact of individual omics layers, establishing practical guidelines for integration within our framework. Overall, our pathway activity autoencoder frameworks yield superior latent representations that are biologically meaningful and are directly translatable into clinically relevant insights.
Code related to this work will be made freely available at \url{github.com/phcavelar/pathwayae} \\
\textbf{Keywords:} Multi-Omic Integration; Autoencoder; Biologically-Informed Neural Networks; Interpretable Deep Learning; Breast Cancer
\end{abstract}

\section{Introduction}

\begin{sidewaysfigure*}[htbp]
	\centering
		\includegraphics[width=.95\textwidth]{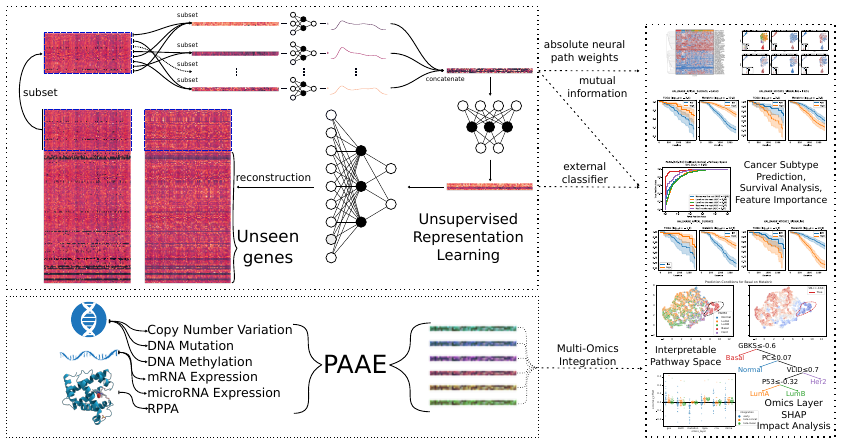}
	\caption{Diagram of the PAAE model. The representation learning module (left) receives a subset of the input gene expression matrix and constrains the architecture to reflect specified pathway gene sets, each producing a single pathway activity value $a$. These activities are further transformed into an internal latent representation $z$, which is then used to reconstruct the entire gene set. Results from downstream tasks are shown on the right. Heatmaps correspond to the best-performing configuration identified for the ``Hallmark Genes'' pathway set. Matrices are drawn to scale and subset for readability. The PAAE framework yields directly interpretable pathway activity scores and higher-level latent representations that can be leveraged for downstream tasks such as classification, clustering, and survival analysis. Each pathway activity is independently interpretable, and its features can be inspected for fine-grained biological insights.}
	\label{fig:diagram}
\end{sidewaysfigure*}

Large-scale analyses of molecular profile data have the potential to transform cancer genomics and are pivotal for the advancement of precision medicine and personalised therapies \citep{bode_precision_2017,chen_moving_2021}. However, the integration of high-dimensional, heterogeneous molecular data, such as genomics, transcriptomics, and proteomics, pose significant challenges, particularly in balancing predictive accuracy with interpretability in machine learning models \citep{selby_visible_2024,selby_beyond_2025}. While deep learning has demonstrated remarkable success in drug response prediction \citep{xie_learning_2024,abbasi_predicting_2024}, phenotypic classification \citep{van_hilten_gennet_2021,yap_verifying_2021}, disease subtype classification \citep{chaudhary_deep_2018,uyar_multi-omics_2021}, and survival prognosis \citep{ching_cox-nnet_2018}, its translational impact is often limited by the inherent opacity of its black-box representations \citep{selby_visible_2024,selby_beyond_2025}.

Interpretability is critical in medical applications, where decision-making must be transparent and biologically plausible \citep{kaur_data_2022}. Post-hoc explainability methods (e.g. \citep{lundberg_unified_2017}, \citep{ribeiro_why_2016}) explain individual features retrospectively, but do not constrain the internal factors in the model and often require re-analysis for each prediction or dataset \citep{li_optimisation-based_2022, dwivedi_explainable_2023, abdrakhimov_prediction_2024}, which limits their utility in clinical applications.
A recent shift has been the incorporation of biological prior knowledge into neural network architectures, giving rise to so-called \textit{Biologically Informed Neural Networks} (BINNs), also called \textit{visible neural networks} (VNNs). These models embed known structures, such as biochemical pathways or gene-sets, into the model architecture, thereby enhancing interpretability and potentially improving generalisability. Recent studies (\citet{selby_visible_2024,selby_beyond_2025}) highlight the ability of BINNs to bridge the gap between predictive performance and mechanistic insight, with models that are explainable by design and do not lag in performance to traditional models.

The models developed recently \citep{da_costa_avelar_pathway_2024} fall inside the class of BINNs that use Biochemical pathways as the main prior-knowledge source for our architectural constraints, which, in this paper, we use to provide an avenue to explore the multi-omic interactions from a pathway-centric perspective. The main contributions of this paper are: 
\begin{enumerate*}[label=(\roman*)]
    \item An update of the methodology for identifying pathways relevant to subtype classification by using OvR Mutual Information, ensuring that even rare subtypes are represented in the selected pathways;
    \item Thorough repeatability and robustness studies, with 2560 models having been evaluated in total showing that while dropout, a common form of regularisation used to promote repeatability, does improve correlation between models, it might also have a negative impact on performance;
    \item A fully automated pipeline to extract a completely interpretable model is provided, resulting in a decision tree with the same number of leaf nodes as there are subtypes;
    \item Prior-knowledge-enhanced multi-omics integration is demonstrated to be beneficial to using single-layers; and
    \item A thorough analysis of the impact of individual omics layers in end-tasks for the TCGA Breast Cancer dataset is performed through an adapted Shapley Value analysis.
\end{enumerate*}

\section{Materials and Methods}

In this section, we briefly outline the PAAE framework in Subsection~\ref{paae-je:ssec:paae}, and highlight recent methodological advances over previous work \citep{da_costa_avelar_pathway_2024}, as illustrated in Figure \ref{fig:diagram}. 
We then detail the repeatability and robustness analyses (Subsection~\ref{paae-je:ssec:repeatability}), describe the multi-omics integration strategies employed (Subsection~\ref{paae-je:ssec:integration}), and present the methodology used for identifying individual-layer contributions to the downstream tasks (Subsection~\ref{paae-je:ssec:omics-effect}). Finally, we summarise the settings used for the downstream tasks in Subsection~\ref{paae-je:ssec:downstream}.

\subsection{The PAAE Framework} \label{paae-je:ssec:paae}

The PAAE framework \citep{da_costa_avelar_incorporating_2023}, shown in Figure~\ref{fig:diagram}, is a Biologically-Informed Neural Network \citep[BINN,][]{selby_beyond_2025} framework based on the Autoencoder \citep[AE,][]{hinton_reducing_2006} and Variational Autoencoder (VAE) deep learning architectures. It works by incorporating pathway information in its architecture, building one pathway encoder block $E_{p}$ for each pathway $p$ in a pathway set $P$. Each of these pathway encoders receive as input $x_{p}$ only the molecular features which map to a gene that belongs inside the pathway $p$, and then output a single pathway activity score $a_{p} = E_{p}(X_{p})$. Thus, we generate a pathway activity space $a$ by concatenating $\|$ all individual pathway activities, as in Equation~\ref{eq:paae-a}

\begin{equation} \label{eq:paae-a}
    a = \left\|_{p\in P}a_{p}\right. = \left\|_{p\in P}E_{p}(x_{p}) \right.
\end{equation}

While in the original formulation, this pathway activity space is further encoded into a latent space $z$, given that we are interested in only in the interpretable pathway activity space, we can consider that $a$ gets further decoded by a decoder $D$, which reproduces the full molecular input $x$. That is, we have a reconstruction $\hat{x} = D(a)$, which is then used to optimise the network parameters for all $E_{p}$, $\Omega$, and for $D$, $\Gamma$ using the a distance function $d$. In practice, we use the mean squared error as $d$, since we normalise our input, giving us the optimisation objective in Equation~\ref{eq:paae-opt}.

\begin{equation} \label{eq:paae-opt}
    \min_{\Omega,\Gamma}\left(\hat{x}-x\right)^{2}
\end{equation}

\subsection{Downstream Tasks} \label{paae-je:ssec:downstream}

For classification, we used the best-performing classifier found during internal validation on the RNAseq study (i.e. Logistic Regression LR), using as input either the latent space $z$ or pathway activity space $a$, to predict the cancer subtype. Model quality is assessed with usual classification metrics (Accuracy, Precision, Recall, and the Area Under the Receiver Operating Characteristic Curve, ROC AUC), macro-averaged and, in the case of ROC AUC, one-vs-rest (OvR) was used. Cancer subtype identification is also performed in an unsupervised fashion through clustering and clustering performance is evaluated using Mutual Information (MI) between the generated clusters and the target classes. Additionally, MI is also used to analyse latent and input features without training a classifier, either by measuring the MI between the feature and the discrete class label (as was done in the RNAseq-only study), or between a feature and a binary indicator variable for OvR analyses.

For survival analyses, we build univariate Cox Proportional Hazard (Cox-PH) models to assess whether individual features are significantly associated with  survival. When a feature is significant, values are stratified into tertiles (low, medium and high), and logrank tests are performed between the low and high groups. In cases where two datasets are compared, the consistency of the direction of survival association (i.e. sign matching between low and high groups) is also taken into consideration. Additionally, we may use the Concordance-Index (CI) of a Cox-PH model \citep{cox_regression_1972} as a ranking metric for survival relevance, analogous to how MI is used for classification tasks. For further details, see Supplementary Material Sec.~\ref{app:sec:survival-analyses}.

\subsection{Repeatability and Robustness Analyses} \label{paae-je:ssec:repeatability}

For robustness and repeatability analyses, we measure the Pearson Correlation between neurons across different repeated random initialisation and training runs, following work reported in  \citet{fortelny_knowledge-primed_2020} and \citet{seninge_vega_2021}. We expand on this work by incorporating other correlation metrics, such as Kendall's $\tau$-b coefficient \citet{kendall_treatment_1945} and Centered Kernel Alignment (CKA, \citet{cristianini_kernel-target_2001,kornblith_similarity_2019}) similarity as a measure of overall repeatability, by analysing both the aggregate ranking of downstream-task metrics and the ranking agreement between models through Average Precision \@ K ($AP@K$ \citep{manning_introduction_2008}, and by performing experiments building aligned consensus models from multiple repeats.

\subsubsection{Aligned Consensus Models}

When performing repeatability and robustness analyses, we calculated Aligned Consensus Models (ACM) in order to study whether they were more repeatable than our non-consensus models. To build an ACM, one takes the pathway activity vector $a_{j}$ of a repeat $j \in [1,\ldots,k]$ out of $k$ repeats. We then take the first model $a_{1}$ as a reference, and calculate the aligned average pathway activity vector $\overline{a}_{i}$ for pathway $i$ as in Equation~\ref{eq:aligned-consensus-model}.

\begin{equation}\label{eq:aligned-consensus-model}
    \overline{a}_{i} = \frac{1}{k} \sum_{j\leq k} \sign(\corr({a}_{j_{i}},{a}_{1_{i}})){a}_{j_{i}}
\end{equation}

That is, we take the average of the pathway activity vector $a_{j_{i}}$ for pathway $i$ and repeat $j$, multiplied by the sign $\sign(c_{{1,j}_{i}})$ of a correlation metric $\corr$ between it and the reference model's pathway activity vector $c_{{1,j}_{i}} = \corr({a}_{j_{i}},{a}_{1_{i}})$. In practice, we take the Pearson Correlation Coefficient for $\corr$.

\subsubsection{Ranking Agreement}

To compute the ranking agreement between two individual repeats $j,m \in [1,k]$, we take a metric $f(a,y)$ that takes a pathway activity vector $a$ and a target $y$, and computes a score $s_i$ for each pathway $i$ w.r.t. to that metric. We then take the two rankings $R_j$ and $R_m$, where $R_j[r] = i$ means that pathway $i$ in repeat $j$ was the $r$-highest score among the pathways. We calculate the Average Precision \@ K ($AP@K$) metric \citep{manning_introduction_2008}, modified to only calculate relevant items up to $K$ as shown in Equation~\ref{eq:acc-at-k}.

\begin{equation}\label{eq:acc-at-k}
    AP@K_{j,m} = \frac{1}{K} \sum_{r<K} P(j) \cdot \mathbb{1}(R_{j}[r] \in R_{m}[:K])
\end{equation}

Where $P(j)$ is the precision at rank $j$, $\mathbb{1}$ is an indicator function for whether the item is in the relevant items, and $R_{m}[:K]$ are all items from the beginning up to $K$ in ranking $R_{m}$. Note that, we assume that $K \leq |P|$.

\subsection{Early and Late Multi-Omics Pathway Activity Inference and Analysis} \label{paae-je:ssec:integration}

Multi-omics integration is categorised into sequential, late and early (joint) integration methods \citep{uyar_multi-omics_2021}. Late integration methods analyse each omics layer independently before combining results, capturing shared patterns across layers. Early integration methods analyse all layers simultaneously, enabling detection of complex inter-omic interactions. In this section, we extend a single-omics framework to multi-omics using both early and late pathway-based integration. We will consider $|O|$ \textit{views} (i.e. Omics layers), where each omics profile $o$ is represented as $x_{o}$, assuming pre-processing has been completed.

In our early integration approach, we combine all omics profiles into one comprehensive profile, $x$ (Equation~\ref{early-input}). Each pathway encoder, $E_{p}$, then specifically focuses on the molecular features linked to the genes ($G_{p}$) that are relevant to its particular pathway ($p \in P$).

\begin{equation} \label{early-input}
    x = \|_{o \in O} x_{o}
\end{equation}

PAAE/PAVAE is optimised to reconstruct the unified profile $x$, enabling each pathway activity score $a_{p}$ to capture complex cross-omics interactions and jointly represent correlated features while reconstructing the full input.


For late integration, we train $|O|$ separate PAAE/PAVAE models with distinct parameters, each producing pathway activity scores $a_{o}$ (Equation~\ref{eq:omic-pathway-activity}). If an omics layer lacks genes for pathway  $p$, we exclude it from that model, denoted as $p \not\in A_{o}$.

\begin{equation} \label{eq:omic-pathway-activity}
    a_{o_{p}} \approx E_{o_{p}}(x), E_{o_{p}}(x) = ({E_{o_{p}}}_1 \cdot \ldots {E_{o_{p}}}_k)(x_{o})
\end{equation}

After generating pathway activities from each model, we integrate using two methods: \textbf{late-mean} integration, and \textbf{late-concat} integration. Late-mean integration takes the arithmetic average of the pathway activities of each model to yield an integrated pathway activity vector $\overline{a}$ (Equation~\ref{eq:late-mean}):

\begin{equation}\label{eq:late-mean}
    \overline{a}_{p} = \frac{1}{\sum_{o \in O, p\in A_{o}}1} \sum_{o \in O, p\in A_{o}} a_{o}
\end{equation}

For late-concat integration, individual pathway activity vectors from each omics layer are concatenated, producing an integrated pathway activity vector $\overline{a}$ (Equation~\ref{eq:late-concat}, where $\|$ denotes concatenation along features).

\begin{equation}\label{eq:late-concat}
    \overline{a} = \|_{o \in O} a_{o}
\end{equation}

\subsection{Individual Omic Layer Contributions} \label{paae-je:ssec:omics-effect}

To evaluate the contribution of each omics layer on downstream performance, we adapt the Shapley value $\varphi_{i}(v)$ \citet{shapley_value_1953}, which approximates a player's average marginal contribution in a cooperative game. 

The definition of the Shapley value $\varphi_{i}(v)$ for a player $i \in N$ in a game with $|N|$ players, is calculated by a weighted average of the value function $v$, measuring the the outcome of the game when a subset players that excludes $i$, $S \subseteq N \setminus \{i\}$ participate, and subtracting it from the value if player $i$ had joined the game, as is shown in Equation~\ref{eq:shapley} below.

\begin{equation} \label{eq:shapley}
    \varphi_{i}(v) = \frac{1}{n} \sum_{S \subseteq N \setminus \{i\}} \binom{n-1}{|S|}^{-1} (v(S \cup\{i\}) - v(S))
\end{equation}

We then adapt this idea to the concept of a multi-view (i.e., multi-omics) machine learning problem, similarly to that in Shapley regression values \citep{lipovetsky_analysis_2001}. Given a set of views (omics layers) $O$ and a performance metric $v$, we calculate the average marginal contribution a view $o$ as a function $f$ of the difference between the contribution of every subset of features $S$ such that $S \subseteq O \setminus \{o\}$, as in Equation~\ref{eq:marginal-contrib}.

\begin{equation} \label{eq:marginal-contrib}
    \varphi_{i}(v) = f(\{(v(S \cup\{o\}) - v(S)) \forall S \subseteq O \setminus \{i\}, S \not= \emptyset\})
\end{equation}

In our specific case, we use $f$ as the median value of all marginal contributions, which we define using a cross-validated metric for $v$, such as ROC AUC or Concordance-Index, meaning we train a model based on view combination $S$ and $S \cup \{o\}$. We ignore the dummy case (i.e. where there is no input, $S=\emptyset$). We also display the distribution of the marginal contributions in plots.

\section{Experimental Setup}

\subsection{Repeatability Studies}

To assess model repeatability, which in this context refers to the ability of our models to produce consistent results when run multiple times under identical conditions on the same dataset, we use the \texttt{MO-TCGA-BRCA} dataset instead of the single-omics TCGA-BRCA dataset, as this work is intended to support downstream multi-omics analyses. All models were trained using the same hyperparameters established during internal validation of the single-omics analysis. To ensure computational efficiency, we restricted our experiments to PAAE models trained on the Hallmark Genes pathway, which exhibited the lowest memory and runtime requirements, thereby enabling more extensive evaluation. We build models for pathway activity dropout values of $Pr(\text{dropout}(a)) \in \{0,0.1,0.2,\ldots0.9\}$, with the highest dropout level resulting in an average of approximately five active pathways per training epoch. For each dropout level, 16 repeats per aligned consensus model were conducted and 16 consensus models were built, yielding $k=256$ models per dropout value, and thus 2560 models in total.

We employ \texttt{scikit-learn} v1.0.2 for mutual information (MI), using \texttt{feature\_selection.mutual\_info\_classif} to estimate the mutual information between pathway activities and PAM50 subtypes,  and \texttt{metrics.mutual\_info\_score} for one-vs-rest (OvR) MI. For survival analysis, we use \texttt{lifelines} v0.27.0, specifically \texttt{CoxPHFitter.concordance\_index\_} to compute the Concordance Index (CI) of individual pathways or models. Correlation metrics are computed using \texttt{scipy} v1.8.1 \texttt{stats.pearsonr} for Pearson's correlation coefficient,  \texttt{stats.kendalltau} for Kendall's $\tau$-b Coefficient, and \texttt{stats.rankdata} to derive pathway rankings $R_{j}$. Ranking agreement is assessed using an internally developed implementation of average precision at rank $AP@K$ (Equation~\ref{eq:acc-at-k}).

\subsection{Interpretable Pathway Activity-Based Subtype Classification}

For our intrepretability study, we use a PAAE model based on the KEGG pathway set previously trained on the single-omics TCGA dataset. Given that, our goal is to propose a highly interpretable model to aid clinical application for subtype classification, we first identify the top pathway for each PAM50 subtype based on the highest OvR Mutual Information, and restrict our feature set to these selected pathways. We then generate UMAP representations of the resulting feature space, and produce featuremaps as visual representations of pathway activity. Subsequently, we train a decision tree classifier on these features, constraining the maximum number of leaf nodes to the number of subtypes. To address class imbalance and ensure that each subtype is represented in the terminal nodes, we apply Synthetic Minority Oversampling Technique (SMOTE) \citep{chawla_smote_2002} prior to training the decision tree.

The decision rules derived by the decision tree nodes are interpreted as logical expressions. These logical conditions are used to subset samples within the UMAP space and identify approximate regions associated with each subtype. For each leaf node, we compute kernel density estimates (KDEs) over the UMAP embeddings by taking the union of all decision path conditions leading to that node.  This approach enables us to visualize the spatial concentration of samples corresponding to specific subtypes in the reduced-dimensional space. We provide these visualisations as well as confusion matrices and ROC curves for our models on both the training (TCGA) and the external validation/test (Metabric) datasets, to demonstrate that this simplified, fully-interpretable model, is capable of achieving reasonable classification performance.

\subsection{Multi-Omics Analyses} \label{sec:exp:sub:mo}

The \texttt{MO-TCGA-BRCA} dataset is employed for multi-omics analysis, and a PAAE model is trained with the same hyperparameters selected in the internal validation of single-omics analysis, with the Hallmark Genes pathway. We train models for every possible combination of the 6 omics layers available (63 combinations in total). External validation is not performed in this setting, as the Metabric dataset lacks several omics layers available in the TCGA dataset and is therefore incompatible with multi-omics integration. To ensure stability in survival analysis, features with variance lower than 0.01 over survival statuses are excluded\footnote{This helps avoid features which completely determine the survival status over the cross validation runs}, and apply an $l^2$ norm of 0.001.

Mapping from gene-based omics layers (such as gene expression, copy number variation, mutation and RPPA) to biological pathways is straightforward. For methylation, we used the CpG site-to-gene mapping provided by the manufacturer for the Illumina HumanMethylation450 BeadChip. For the $\mu$RNA layer, we used the miRTarBase2025 \texttt{miRTarBase\_SE\_WR} database of $\mu$RNA target genes which includes target gene interactions supported by strong experimental evidence \citep{cui_mirtarbase_2025}. Alternative strategies were considered and described in Supplementary Material \ref{app:sec:omics-mapping}.

\section{Results and Discussion}

\subsection{Feature Space Visualization for Intepretability Analysis}

\begin{figure*}[htbp]
    \centering
    \subcaptionbox{TCGA (train)\label{fig:je-clustermap:sub:tcga}}{\includegraphics[width=.45\linewidth]{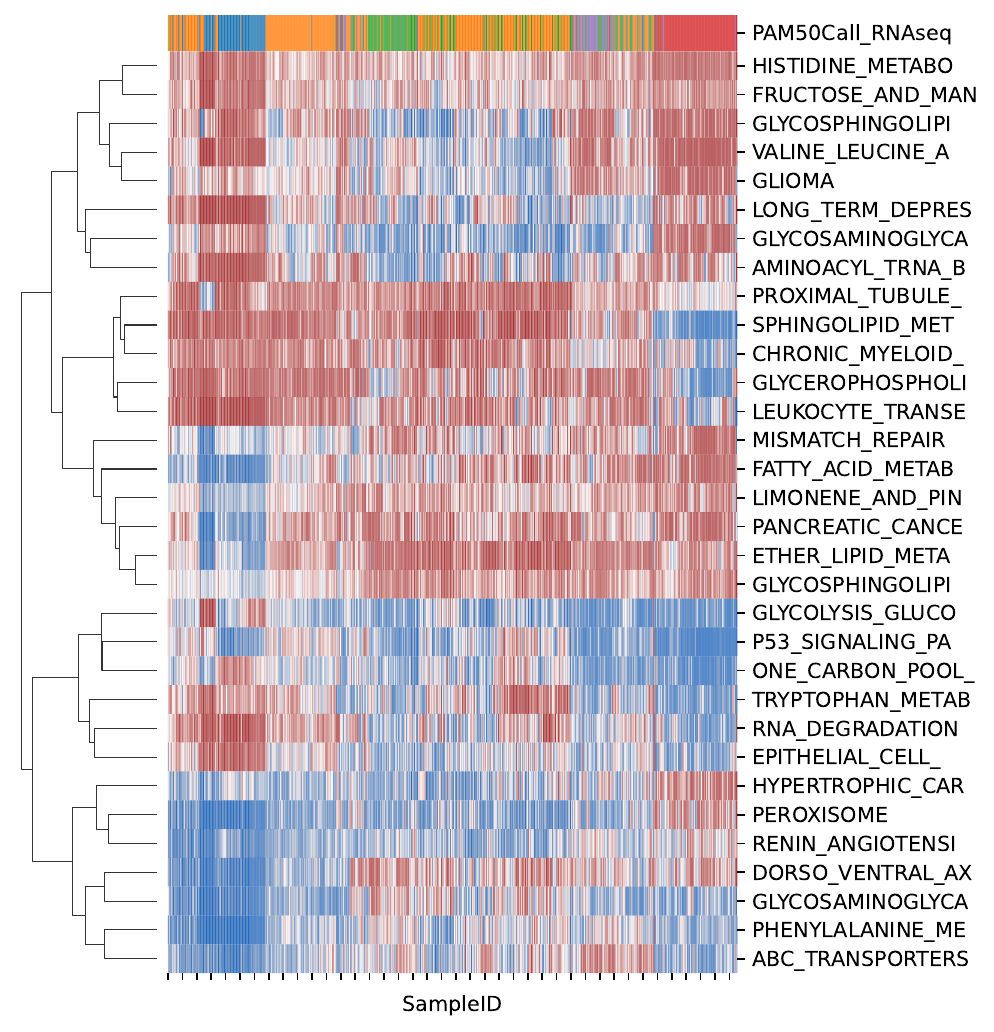}}
    \hfill
    \subcaptionbox{Metabric (test)\label{fig:je-clustermap:sub:meta}}{\includegraphics[width=.45\linewidth]{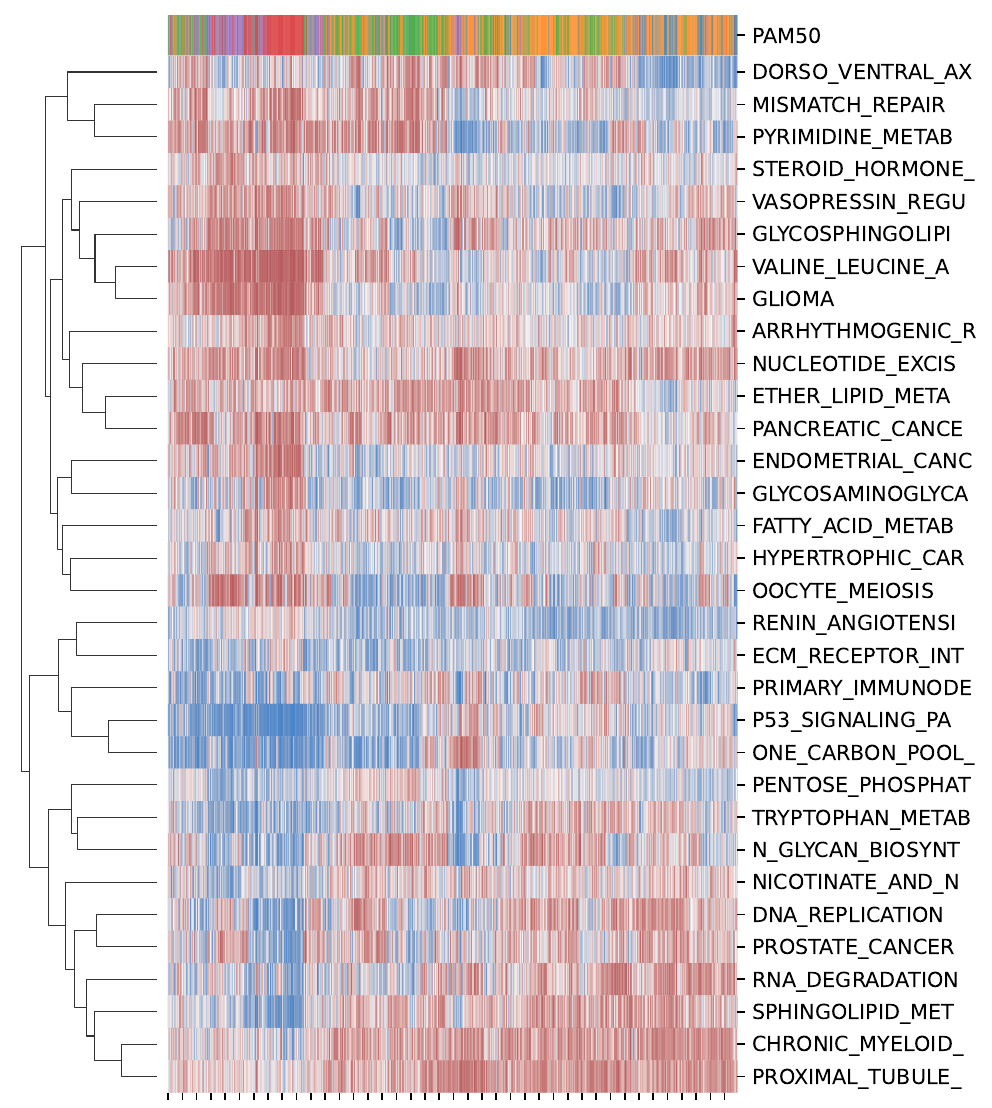}}
    \caption{Clustermap based on cosine distances between sample-level inferred pathway activity vectors in the KEGG PAAE pathway activity space (see Sec.~\ref{app:sec:clustermaps}). Colours denote BRCA clinical phenotypes ({\color{brca-normal}Normal in blue}, {\color{brca-luma}Luminal A in orange}, {\color{brca-lumb}Luminal B in green}, {\color{brca-basal}Basal in red}, {\color{brca-her2}Her2 in purple}). For clarity, only the 32 pathways with highest mutual information with respect to the class labels are shown (see Supplementary Material Sec.~\ref{app:sec:mutualinfo} for details). Fig.~\ref{fig:clustermap-hallmark} shows similar results for the Hallmark Genes pathway set across all 50 pathways.
    }
    \label{fig:je-clustermap}
\end{figure*}

\begin{figure*}[hbtp]
    \centering
    \includegraphics[width=.95\linewidth]{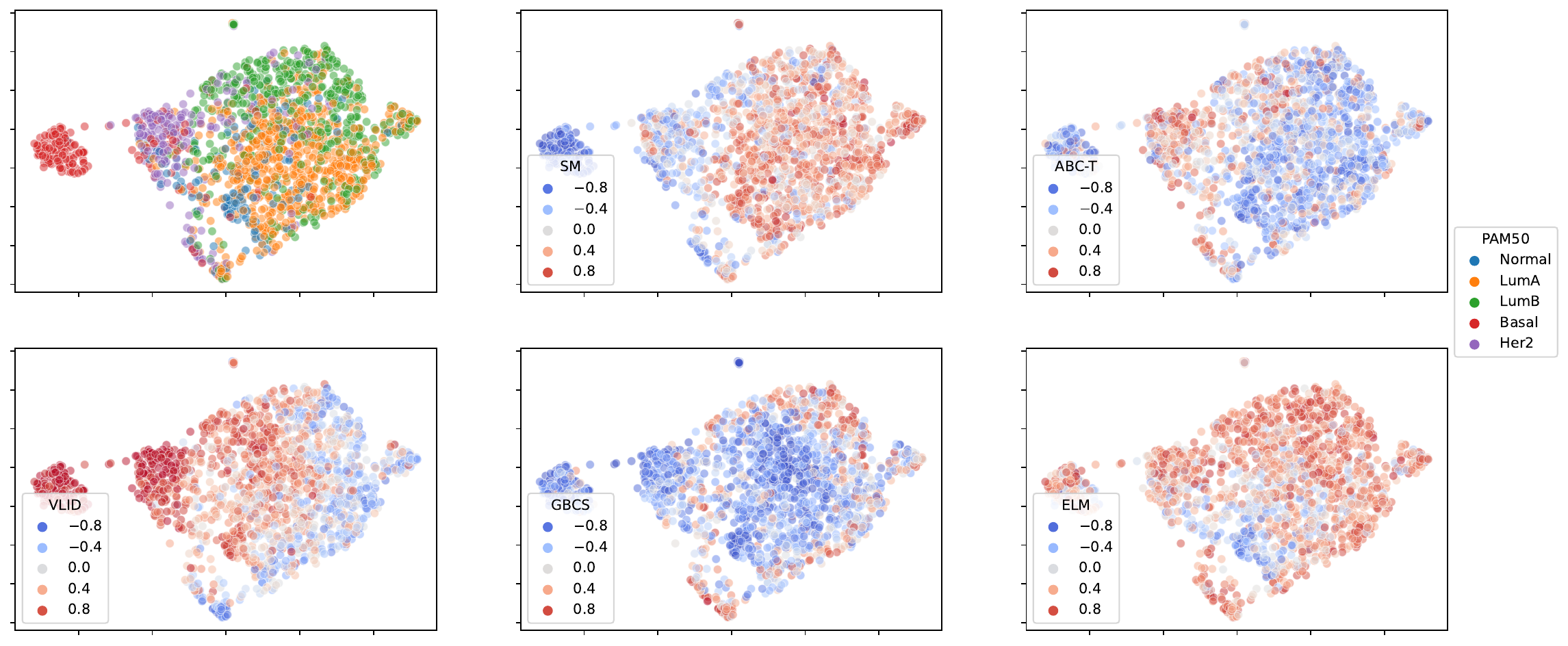}
    \caption{Subtype distribution overlaid on 2D UMAP projection of the inferred pathway activity vectors (top-left), as well as featuremaps showing the intensity of each sample inferred pathway activity scores for KEGG-based PAAE pathway activity space in the Metabric dataset. For clarity, only the 5 pathways with the highest mutual information in TCGA subtypes are shown.}
    \label{fig:je-featuremap}
\end{figure*}

With respect to interpretability, pathway activity scores enable comparisons across subtypes, supporting clinically relevant insights (see discussion in Supplementary Material Section~\ref{app:sec:interpret-pa}). As shown in Fig.~\ref{fig:je-clustermap}, clustering of pathway activity scores with cosine distance yields clear separation aligned with clinical labels, demonstrating that augmenting a model with pathway prior knowledge enhances both interpretability and predictive power, even in an unsupervised setting. Furthermore, mutual information analysis shows that most of the 32 pathways identified as most informative in the training set remain informative in the test dataset, highlighting the robustness of the learned representations.

Using the KEGG pathway set, the top 5 pathways with highest one-vs-rest mutual information for each PAM50 subtype (Basal, Her2, Luminal A, Luminal B, and Normal) in the TCGA dataset were Sphingolipid Metabolism (SM), ABC Transporters (ABC-T), Valine Leucine and Isoleucine Degradation (VLID/VLI DEG), Glycosaminoglycan Biosynthesis Chondroitin Sulfate (GBCS) and Ether Lipid Metabolism (ELM). In the Metabric dataset, Tryptophan Metabolism, P53 Signaling Pathway, Glutathione Metabolism, and Mismatch Repair appeared among the top pathways instead of ABC-T, VLID, GBCS, and ELM. However, the TCGA pathways remained highly ranked at 8th, 3rd, 19th and 2nd place out of 186 total pathways, respectively, in the Metabric dataset indicating strong cross-cohort consistency. We show the feature map of this representation as in the 2D UMAP projection of our pathway activity space, as shown in Figure~\ref{fig:je-featuremap}.

Key pathways prioritised by our methodology, i.e. 
Sphingolipid Metabolism \citep{ryland_dysregulation_2011,corsetto_critical_2023},
ABC-Transporters \citep{xiang_abcg2_2011,muriithi_abc_2020},
GBCS \citep{huang_novel_2021,yen_targeting_2024},
and ELM pathways \citep{benjamin_ether_2013,yu_systematic_2021}, have been documented in tumorigenesis and cancer progression, either specifically in breast cancer \citep{xiang_abcg2_2011,ryland_dysregulation_2011,yu_systematic_2021,corsetto_critical_2023,yen_targeting_2024}, or more broadly across various cancer types \citep{ryland_dysregulation_2011,huang_novel_2021,benjamin_ether_2013,muriithi_abc_2020}. Importantly, we note that the identified Valine Leucine and Isoleucine Degradation (VLID) pathway is relatively under-explored in the context of cancer biology.
Recent studies \citet{zeleznik_branched-chain_2021} report a significant association between elevated circulating branched-chain amino acid (BCAA) levels and lower breast cancer risk in the NHSII cohort. This observation suggests that reduced BCAA catabolic activity may be linked to disease progression. The identification of the VLID pathway as a discriminative feature in our model could reflect underlying metabolic shifts, highlighting its potential as a novel biomarker for cancer prognosis. 

Similarly, recent literature underscores the key role of tryptophan metabolism in breast cancer progression and immune evasion through the immunosuppressive kynurenine pathway \citep{girithar_involvement_2023,ma_global_2025}. Single-cell transcriptomic analyses show macrophages in breast cancer tissues exhibiting elevated tryptophan metabolic activity, which correlates with M1 type polarisation and cytolytic CD8\textsuperscript{+} T cell infiltration, thereby serving as both an immunotherapy response predictor and a marker of subtype specific immune microenvironments \citep{xue_tryptophan_2023}. Taken together, apart from well-studied pathways in cancer (e.g. p53 or mismatch repair), our study highlights pathways of emerging importance (i.e. metabolic and transport pathways above) that act as key drivers of tumour biology. These pathways play crucial roles in resistance mechanisms and immunosuppressive networks in breast cancer, making them promising targets for future research and therapeutic intervention.

\begin{figure*}[htpb]
    \centering 
    \subcaptionbox{\label{fig:je-survival-kegg-importantpathways:vlid}}{\includegraphics[width=.45\linewidth]{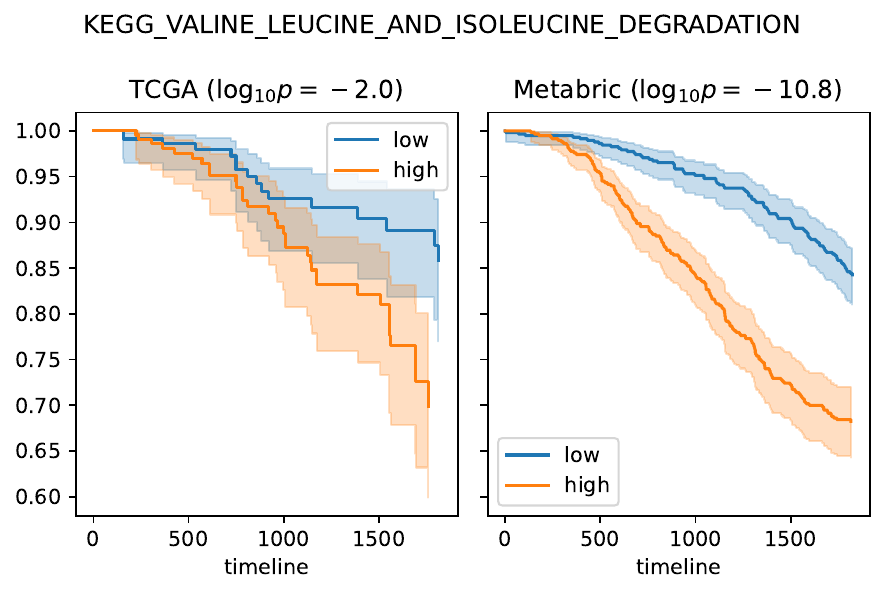}}
    \hfill
    \subcaptionbox{\label{fig:je-survival-kegg-importantpathways:sm}}{\includegraphics[width=.45\linewidth]{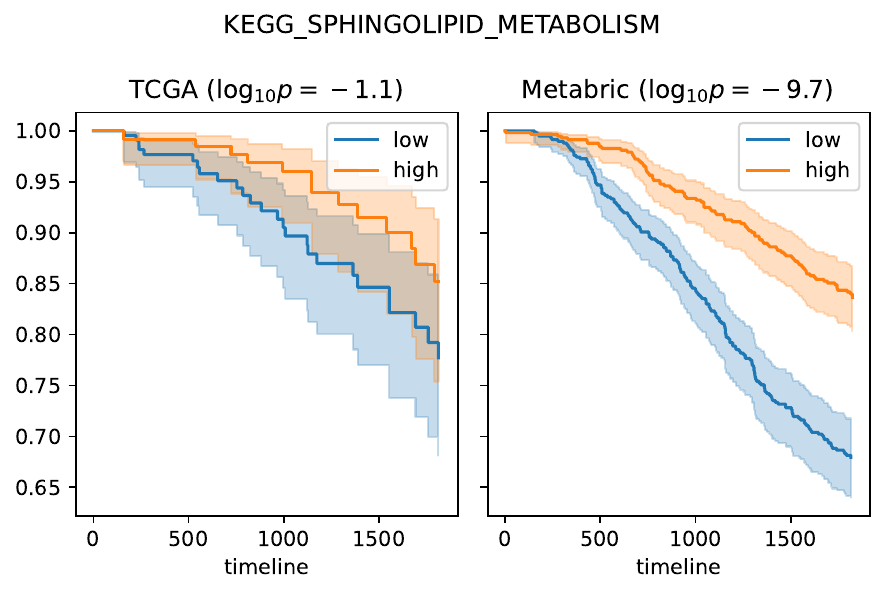}} \\
    \subcaptionbox{\label{fig:je-survival-kegg-importantgenes:sm:degs2}}{\includegraphics[width=.45\linewidth]{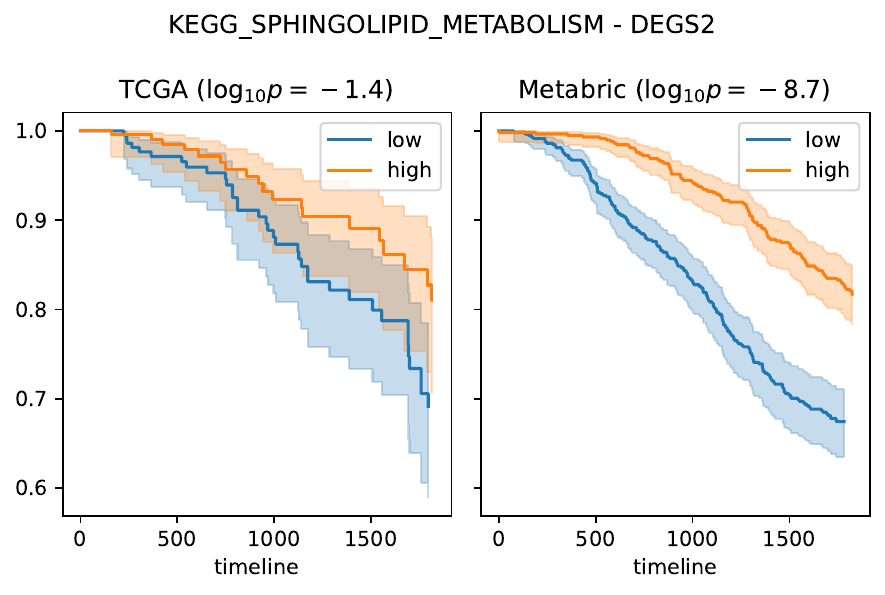}}
    \hfill
    \subcaptionbox{\label{fig:je-survival-kegg-importantgenes:vlid:mccc1}}{\includegraphics[width=.45\linewidth]{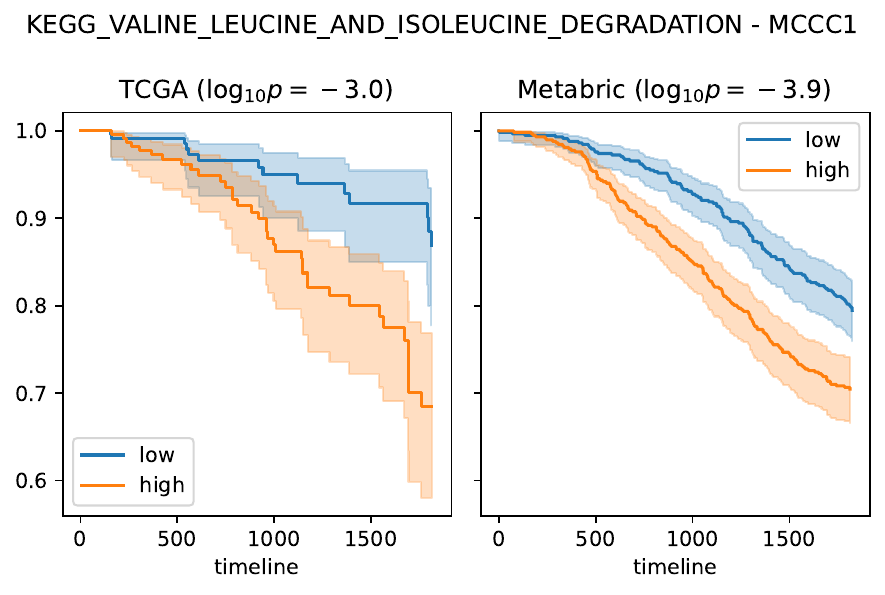}}\\
    \subcaptionbox{\label{fig:je-survival-kegg-importantgenes:vlid:oxct1}}{\includegraphics[width=.45\linewidth]{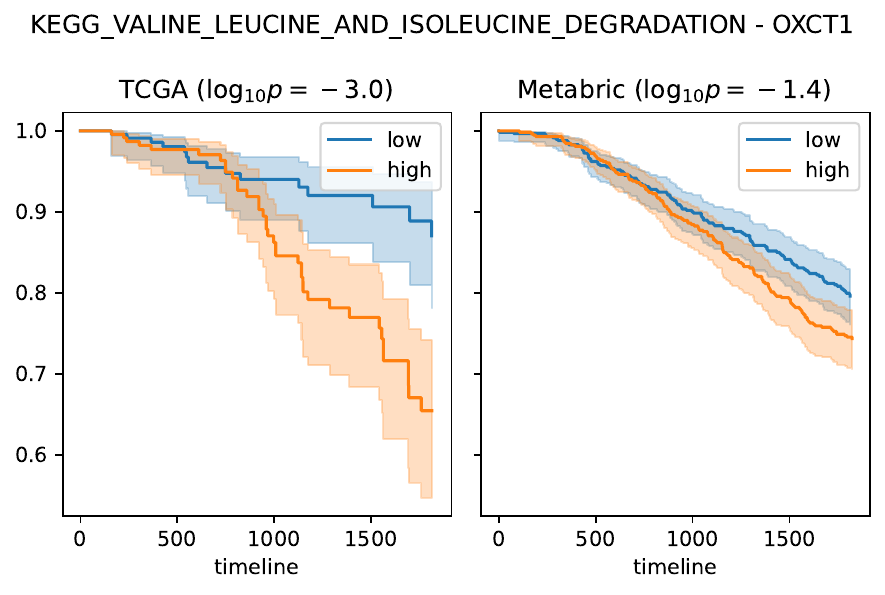}}
    
    \caption{Kaplan-Meier Curves for the pathways (\ref{fig:je-survival-kegg-importantpathways:vlid},\ref{fig:je-survival-kegg-importantpathways:sm}) and genes (\ref{fig:je-survival-kegg-importantgenes:sm:degs2}-\ref{fig:je-survival-kegg-importantgenes:vlid:oxct1}) that showed statistically significant differences (logrank test) between the upper and lower tertiles in expression in both the TCGA and Metabric datasets,  with consistent survival directionality (see Supplementary Material Sec.~\ref{app:sec:survival-analyses} for details). These 4 genes represent $57.14\%$ of the 7 genes that were significant in the TCGA logrank test, drawn from the 10 most informative genes (by OvR mutual information with the classification target) within each of the 5 most relevant pathways.}
    \label{fig:je-survival-kegg-importantgenes}
\end{figure*}

\subsection{Survival Analysis}

For each of above-mentioned pathways (SM, ABC-T, VLID, GBCS, and ELM), 10 genes were identified by highest Absolute Neural Path Weight (ANPW) ranking \citep{uyar_multi-omics_2021} and statistical tests for survival analyses on all 5 pathways and these genes were performed, selecting those with both a significant logrank separation and concordant survival directionality between low and high expression groups at the cutoff date (See Supplementary Material Section~\ref{app:sec:feat-importance} for details). Fig.~\ref{fig:je-survival-kegg-importantgenes} shows the Kaplan-Meier curves for the upper and lower third percentiles of pathway activity scores or TPM/IPM expression values, along with logrank test p-values. It is noted that the VLID pathway shows clear survival stratification, particularly in the Metabric dataset.

\subsection{Ablation study on the effect of dropout on pathway activity models}

\begin{figure*}
    \centering
    \includegraphics[width=0.95\linewidth]{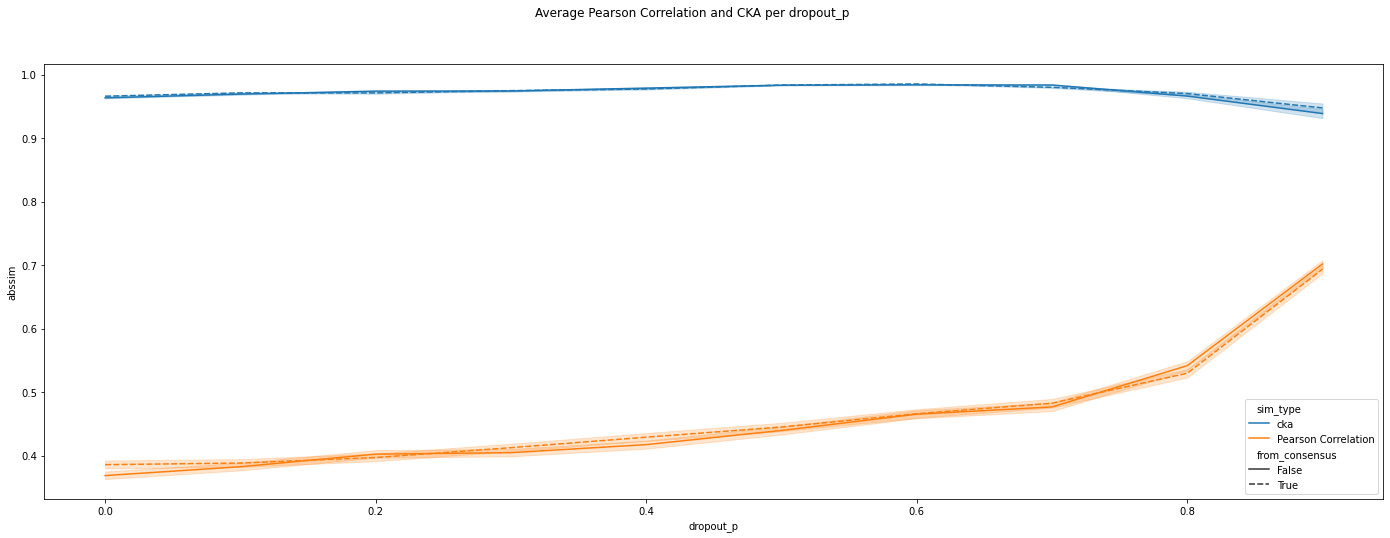}
    \caption{Plot illustrating the average Centered Kernel Alignment (CKA) similarity and the average Pearson Correlation between pathways obtained from models trained with different random seeds on the \texttt{MO-TCGA-BRCA} dataset. Models are labelled as either Aligned Consensus Model or not. Correlation increases as dropout rate increases, with sharp increase between 70 and 80\% dropout, and an accompanying decrease in CKA similarity, indicating that while pathway-specific features become more consistent across models, global representational similarity slightly declines. Shaded areas indicate variance across model runs.}
    \label{fig:dropout-cka-pearsonr}
\end{figure*}

Previous studies in linear BINNs showed improved robustness when applying dropout, specifically in the latent layer \citep{fortelny_knowledge-primed_2020,seninge_vega_2021,selby_visible_2024}. Here, we extend this investigation to the case of a non-linear Visible Neural Network in the form of our PAAE model. As shown in Figure~\ref{fig:dropout-cka-pearsonr}, the Centered Kernel Alignment (CKA) representation is stable across a range of dropout values. Additionally, the absolute Pearson correlation between individual pathway activity vectors across independently trained models increases as the dropout rate also increases, showing that applying dropout on the latent and, in our case, in the pathway activity layer has a positive impact on robustness. Interestingly, a phase transition was observed between 70\% and 80\% dropout, where the Pearson correlation increases sharply, suggesting a shift in model behaviour.

Further investigation revealed that, although higher dropout resulted in models becoming more repeateable, representational power declined, especially beyond the previously mentioned phase transition, as seen in Figure~\ref{fig:dropout-performance} for ROC AUC and Concordance Index.
Heatmaps to visualise models with different dropout rates (Figure~\ref{fig:dropout-heatmap-order}), illustrate that, although increasing dropout makes features more reproducible, it unsurprisingly also ``bleaches'' the signal from pathways, generating pathways that are not only more repeatable across different training runs, but also more similar among themselves. This likely reflects a higher noise-to-signal ratio during the training process, which drives the model toward more generic representations.

Figures~\ref{fig:dropout-kendall-tau-mi}~and~\ref{fig:dropout-kendall-tau-ci} suggest that representations generated beyond the phase transition are significantly different from those at lower dropout. Models with 0-70\% dropout exhibit comparable similarity to those at 50\% dropout, while being markedly different from 90\% dropout. Additionally, the average signal each individual pathway has with respect to survival and classification changes depending on the amount of dropout used especially on the 80\% and 90\% dropout models (see Figures~\ref{fig:dropout-bothmi}~and~\ref{fig:dropout-bothci}). There is an especially marked difference with respect to the Mutual Information with the PAM50 subtypes.  When analysing the ranking agreement across models,  higher dropout seems to reduce consistency in mutual information (Figure~\ref{fig:dropout-rankingmi}), while agreement in survival ranking remains stable, with a marked increase for the highest dropout value (Figure~\ref{fig:dropout-rankingci}). 

\subsection{Interpretable Subtype Classification using the Pathway Activity Space: A Case Study}

\begin{figure*}
    \centering
    \includegraphics[width=0.95\linewidth]{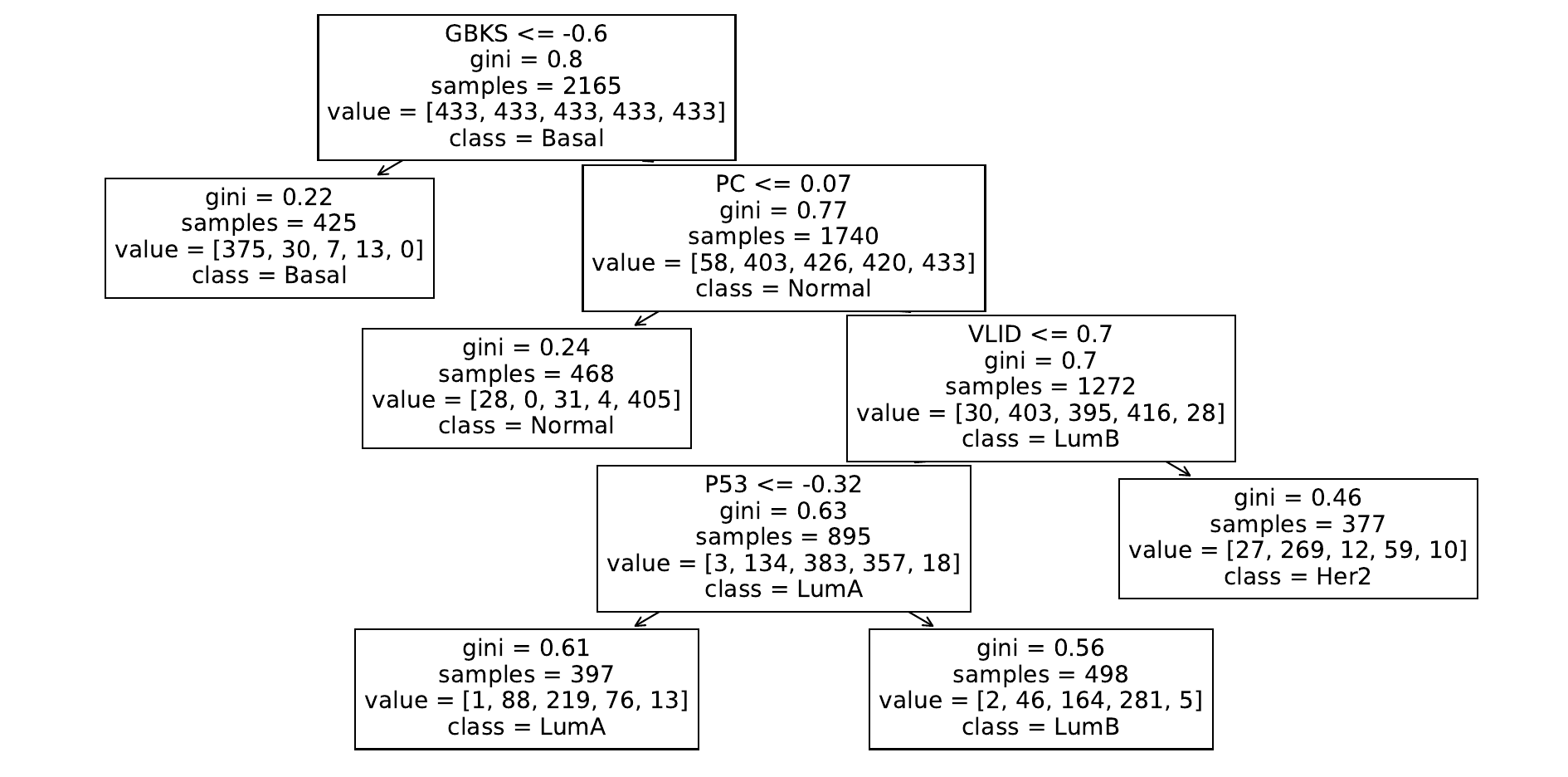}\\
    \caption{Decision tree generated after training it on pathway activity vectors of the PAAE KEGG model trained on the single-omics TCGA dataset. The tree is constrained to 5 leaf nodes, and SMOTE oversampling\citep{chawla_smote_2002} is used to ensure that each PAM50 subtype is represented in the leaf nodes.}
    \label{fig:dtree}
\end{figure*}

Model interpretability tools can assist medical practitioners in understanding and contextualising predictions. However, inherently interpretable models, such as BINNs, enable the generation of direct visualisations owing to the internal representations generated by the model itself.
In this subsection we report a case study demonstrating a fully-interpretable subtype prediction model. Importantly, the objective here is not to optimise predictive accuracy, but to illustrate the interpretability of our model. For this analysis, we use a KEGG-based PAAE model, but similar interpretability techniques could also be applied to other models. The top most informative pathways with respect to Mutual Information for each subtype in the TCGA dataset are selected, and our model is subset to include only these pathways. A decision tree classifier is trained with the maximum number of nodes equal to the number of subtypes. To ensure that all classes are represented in the leaf nodes, SMOTE oversampling was used \citep{chawla_smote_2002}. The resulting decision tree is shown in Figure~\ref{fig:dtree}.

\begin{figure*}
    \centering
    \includegraphics[width=0.95\linewidth]{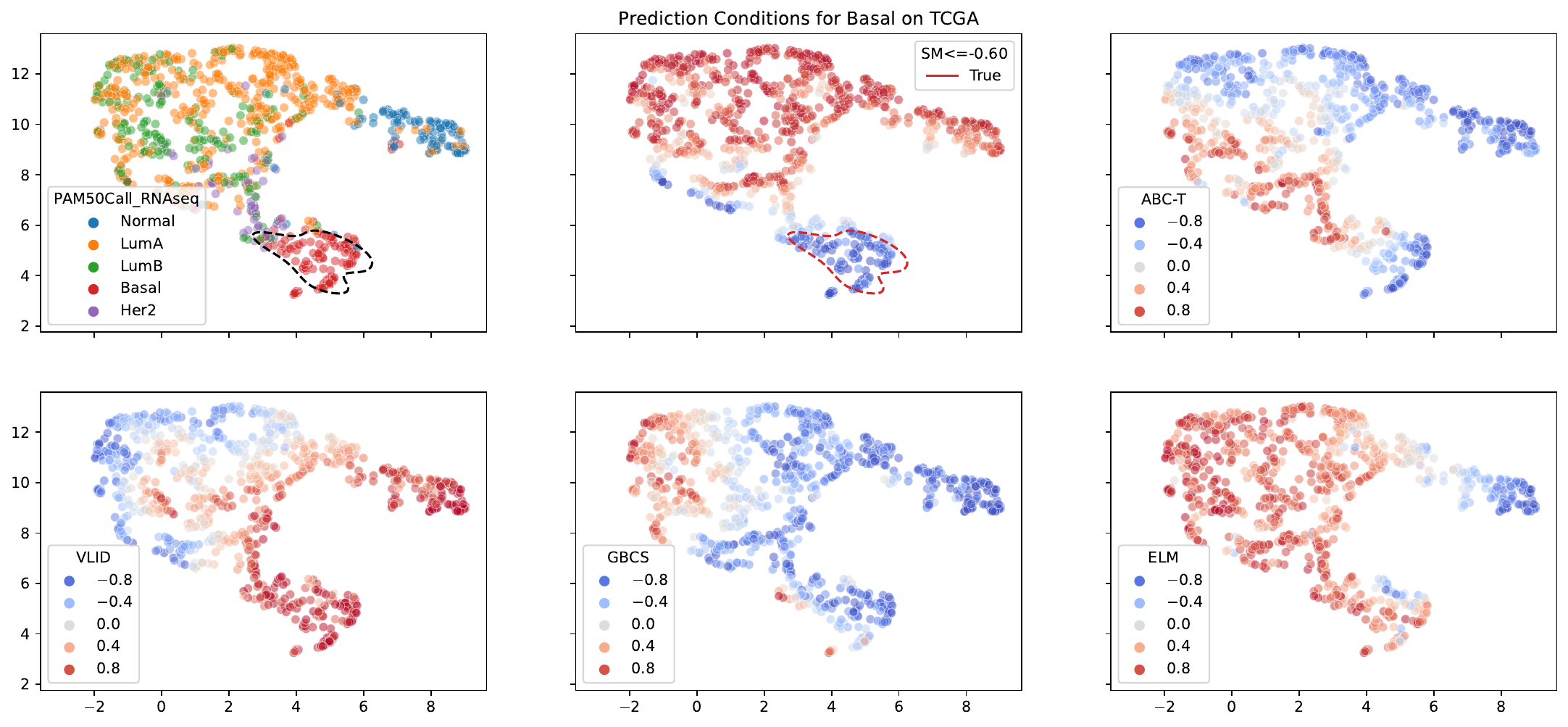} \\
    \includegraphics[width=0.95\linewidth]{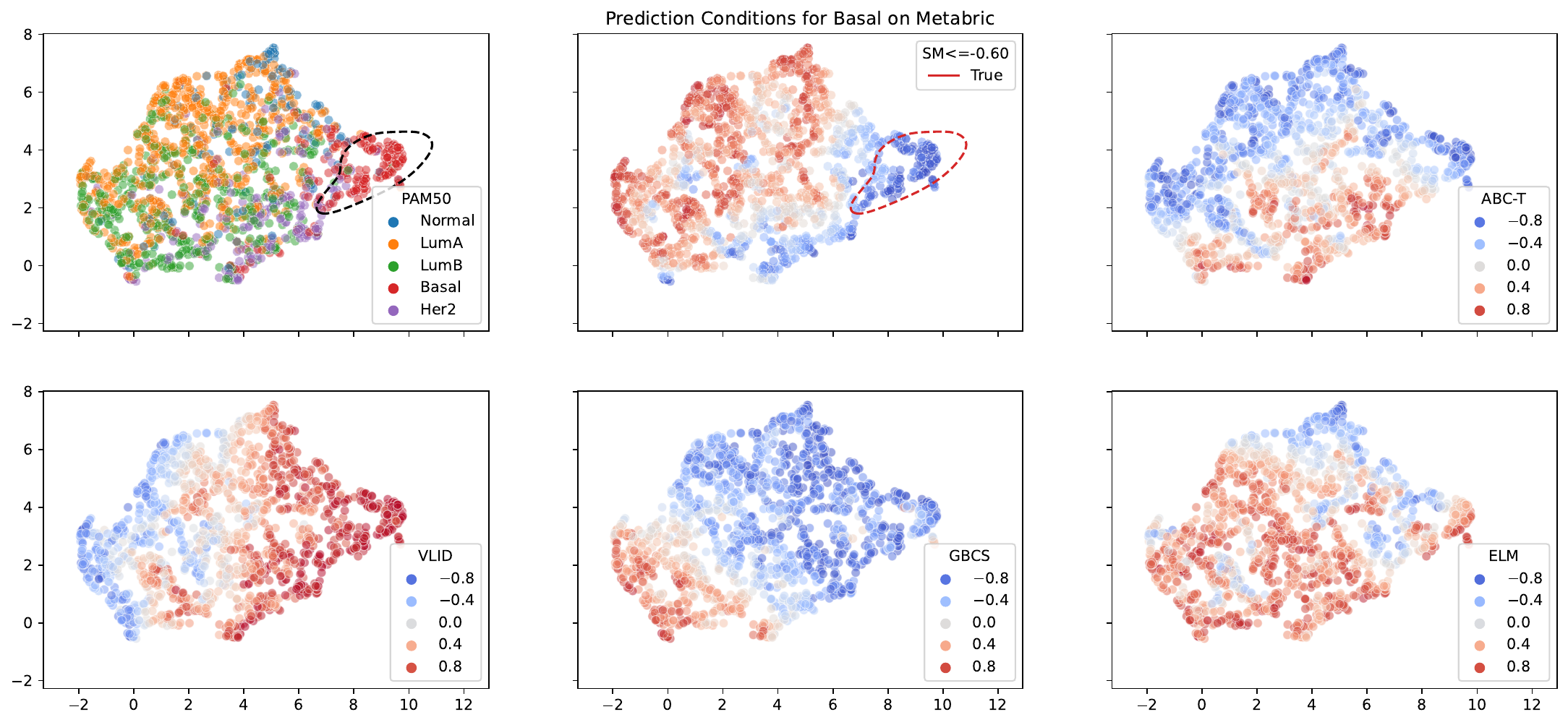}
    \caption{2D UMAP featuremaps for TCGA (top six) and Metabric (bottom six) datasets for pathway activity vectors derived from  the top 5 pathways with highest one-vs-rest mutual information. Red KDE contours are overlaid on the SM pathway featuremap in both datasets, highlighting regions that satisfy the decisions rules for the Basal subtype from decision tree depicted in Figure~\ref{fig:dtree}. For clarity, the same contours are displayed in black on the Class featuremaps.}
    \label{fig:dtree-path}
\end{figure*}

This approach yields a clear, human-interpretable model that separates samples into each of the five subtypes. Furthermore, by adapting the UMAP space to use only the 5 top pathways rather than the full pathway set, a smoother UMAP featuremap in terms of pathway activities that drive these subtypes, is generated. 
In Figure~\ref{fig:umap-full-vs-ovrmi} (top four plots) illustrates UMAP embeddings generated for all KEGG pathways, as well as for the top 5 pathways. In Figure~\ref{fig:dtree-path} two sets of six plots are shown, analogous to those in Figure~\ref{fig:je-featuremap}, showing featuremaps and KDE contours showing the area over the UMAP space. These indicate regions where samples that satisfy the decision tree criteria for classification as Basal in both the TCGA (middle) and Metabric (bottom) datasets. As expected, a model that is constrained to this level of interpretability performs worse than full models, as can be seen from confusion matrices and receiver operating characteristic curves in Figure~\ref{fig:dtree-validation}. While this case study demonstrates the potential for direct interpretability in our approach, it also highlights the complexity of cancer and indicates that accurate and robust prediction requires the integration of a larger set of interacting pathways.

\subsection{Multi-Omics for PAM50 Classification and Survival Prediction}

\begin{figure*}
    \centering
    \includegraphics[width=.99\linewidth]{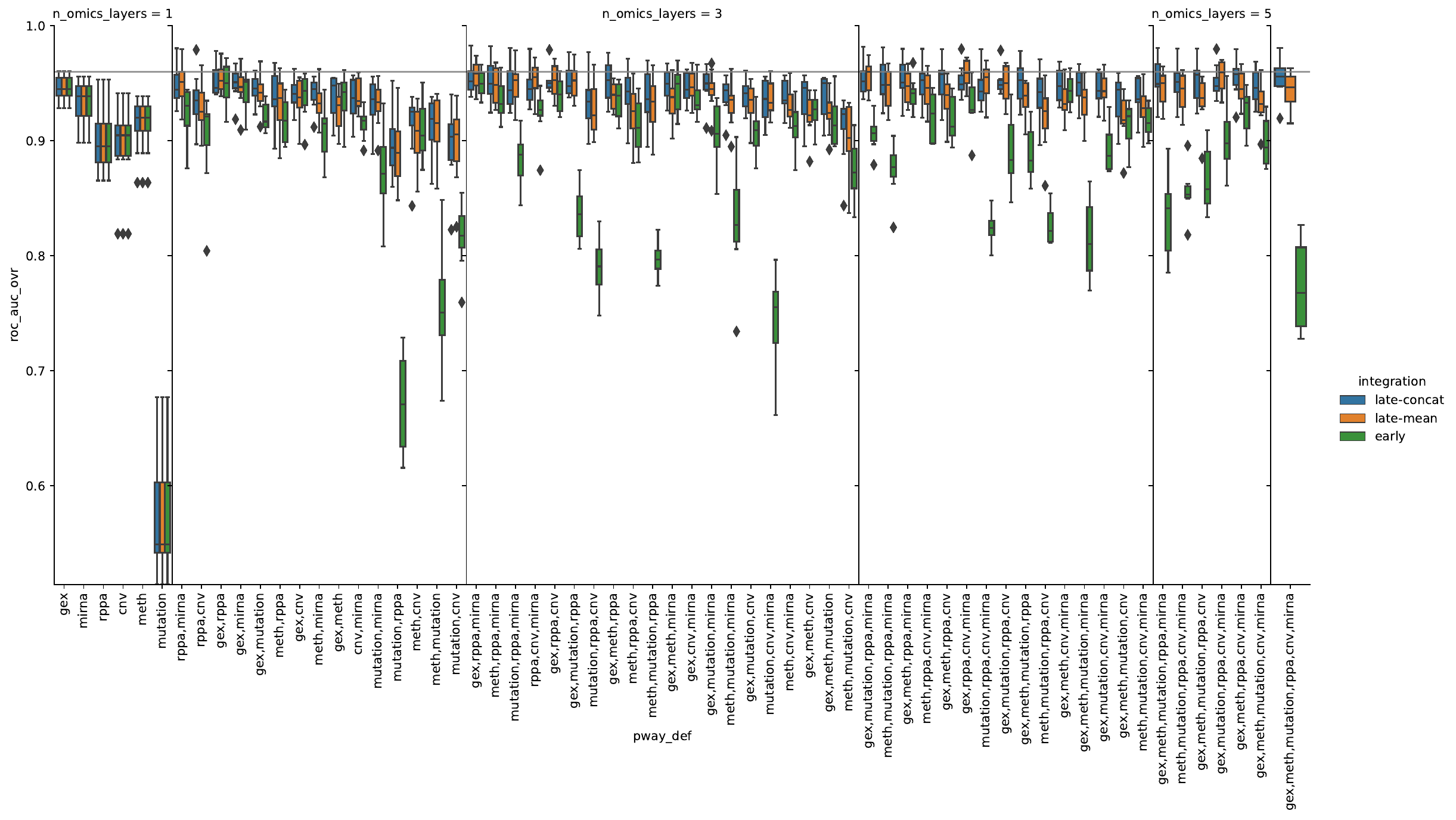}
    \caption{ROC AUC of one-vs-rest logistic regression for all combinations of omics layer in early, late-concat and late-mean integration. The gray line denotes the highest median ROC AUC observed across all configurations tested.}
    \label{fig:clf-roc}
\end{figure*}

This subsection details our multi-omics integration analysis using the PAAE model. Our primary objectives are to demonstrate how multi-omics data can be integrated with PAAE, to assess the general benefits of using multi-omics approaches over single-omics, and to quantify the individual contribution of each omics layer to the integrated model's performance. We illustrate the effectiveness of early versus late integration strategies, as defined in Subsection~\ref{paae-je:ssec:integration}, for multimodal data in the context of PAM50 subtype classification and survival prediction, and determine the effect of specific modalities in data integration using the PAAE method. As shown in Figure~\ref{fig:clf-roc}, early integration rarely outperformed either of the late integration methods for the classification of PAM50, achieving superior performance in only 6 of 63 cases. In most cases, the inclusion of mutation data led to a marked decrease in performance, suggesting that this layer may interact antagonistically with others in the early integration framework. It should be noted that this adverse effect was not observed in the two late integration tests.

The highest median ROC AUC was achieved using late-mean integration of all omics layers except methylation. This was closely followed by late-concat integration of all layers except mutation and of all layers combined. Although other individual layers generally performed worse than gene expression alone, the multi-omics integration approach yielded synergistic improvements in performance, particularly through late-integration approaches.

We also evaluated additional metrics, including classification accuracy (Figure~\ref{fig:clf-acc}), and performed unsupervised clustering using the K-Means algorithm, assessing performance with mutual information (Figure~\ref{fig:cls-mi}) and Rand Index (in Figure~\ref{fig:cls-ri}). The  overall conclusion, i.e. that multi-Omics integration was beneficial, was consistent across these metrics. Although the best-performing methods varied, the best performing models were observed in integrating 4-6 omics layers.

\begin{figure*}
    \centering
    \begin{minipage}{0.99\linewidth}
    \begin{tikzpicture}
      \node (img)  {\includegraphics[width=.99\linewidth]{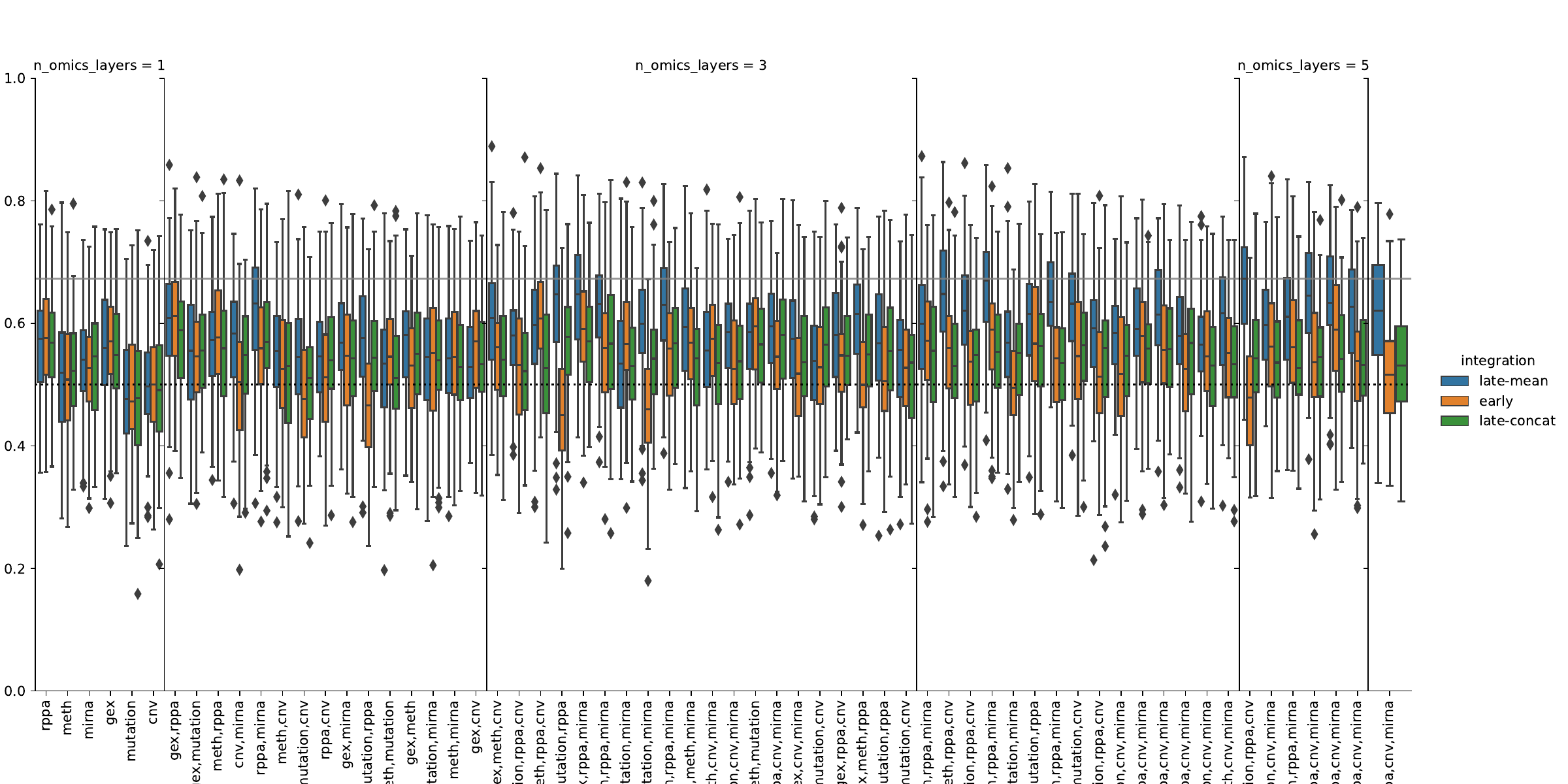}};
      \node[rotate=90, anchor=center, yshift=.5\linewidth, font=\scalefont{0.001}] {concordance\_index};
     \end{tikzpicture}
    \end{minipage}%
    \caption{Concordance Index (C-index) of a Cox proportional hazards model for all omics layer combinations across for three tested integration methods, i.e. early, late-concat and late-mean. The gray line indicates the highest median C-index among all tested configurations (dotted line marks concordance of 0.5).}
    \label{fig:sur-ci}
\end{figure*}

For survival prediction our results also indicate that multi-omics integration can enhance performance (Figure~\ref{fig:sur-ci}). However, the late-concat approach was less consistently effective in this setting, and the negative contribution of the mutation and CNV layers was more pronounced, particularly given that these layers performed worse than random chance.

\subsection{Omics Layer Contributions}

\begin{figure*}
    \centering
    \includegraphics[width=0.99\linewidth]{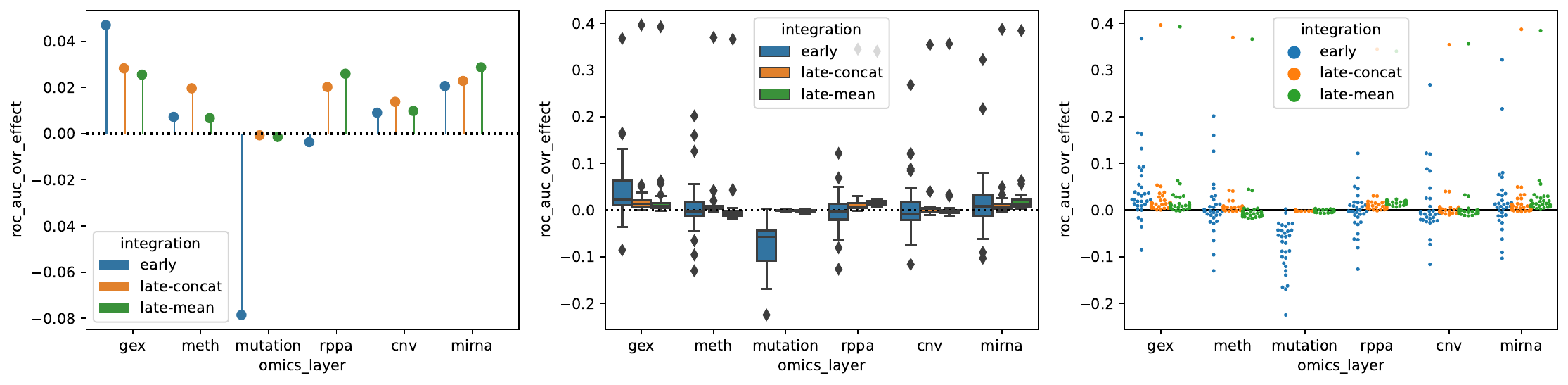}
    \caption{Marginal contribution of each omics layer to the classification performance, measured by ROC AUC of a one-vs-rest logistic regression model, across the three tested integration methods: early, late-concat and late-mean. The left panel shows the median marginal contribution, the right panel displays a swarm plot of values across all combinations excluding that layer, and in the middle the same information is shown as a boxplot.}
    \label{fig:omics-effect-clf-roc}
\end{figure*}

We then analysed the marginal contribution of each omic layer on a previously existing combination, as described in Subsection~\ref{paae-je:ssec:omics-effect}, and shown in Figures~\ref{fig:omics-effect-clf-roc}~and~\ref{fig:omics-effect-sur-ci}, to investigate which omics layer had the greatest individual impact on each task.
The mutation layer consistently exhibited a strong negative impact on  PAM50 classification (ROC AUC) when using early integration (Figure~\ref{fig:omics-effect-clf-roc}). This effect, however, was largely mitigated with late-concat integration, likely because this method allows the classifier to effectively ignore less informative layers. Late-mean integration also reduced the negative influence of the mutation layer. As expected, the gene expression layer was the only one to consistently improve PAM50 classification, likely  because the PAM50 molecular subtypes were based on gene expression profiles.

The $\mu$RNA and RPPA layers showed the next highest  median performance increase, with the RPPA layer exhibiting a positive median effect only in late integration. These layers consistently contributed positively only when late integration was used. We hypothesise that this is due to these integration methods, especially late-concat, allowing noisy pathways within these layers to be ignored or mitigates the effect of these pathways.

As shown in Figure~\ref{fig:omics-effect-sur-ci}, for survival prognostication, the late-mean integration method consistently outperformed all others, and was the only integration method that allowed for an consistently positive impact across all omics layers. The most beneficial layers, in order of median impact, were RPPA, gene expression and $\mu$RNA.

Interestingly, late-concat integration demonstrated worse performance than late-mean integration in this context. While this may appear to contradict our previous observation that late-concat can mitigate the impact of noisy features, further analysis suggests that one of the issues with survival information also comes from how noisy labels were, and how often features seemed to dictate entirely the survival status of a patient, which degraded, and sometimes impeded, convergence.

\section{Discussion}

In this work, we further evaluate our PAAE \citep{da_costa_avelar_pathway_2024} deep learning framework that incorporates prior biological knowledge, with a view to improving patient stratification and prognostic prediction via analysis of multi-omics data, repeatability studies, and by providing changes to our validation. 
Our results reinforce that our pathway activity autoencoder frameworks, although focusing on unsupervised dimensionality reduction, translate into better latent representations, while also providing an intermediate latent representation that is directly translatable into a concept relevant to medical practitioners.

Our analysis of the effect of dropout on our BINN model (PAAE) confirms findings by \citet{fortelny_knowledge-primed_2020} and \citet{seninge_vega_2021}, showing increased correlation between outputs of independently trained models with higher dropout rates. However, our results further demonstrate that excessive dropout can lead to a collapse in model performance, particularly in non-linear BINNs. In other words, although representations become more similar, they exhibit reduced predictive power for both downstream tasks. Furthermore, we observe that dropout not only affects overall performance, but also alters which pathways contribute in predictive tasks, thereby caution in the use of this technique is advisable. We also report an extreme example of interpretability with our model, where we select the pathway that has the most OvR mutual information per subtype to build an extremely simplified model to aid clinical interpretability. We show how to build a decision tree that classifies samples with minimal redundancy in its leaves, and analyse performance, while also providing visualisation tools to aid clinicians in deriving specific pathway combinations that contribute to distinct patient subtypes.

Importantly, our framework works in multi-omic data analysis, and provides key insights into contributions of specific modalities in subtype classification. Our study showed that, although Gene Expression, RPPA and $\mu$RNA layers had the most positive marginal contribution overall, for none of the tasks, a combination composed of exclusively these three layers was the best, and more often than not adding other layers acted synergistically. Furthermore, there seemed to be a disconnect between PAM50 classification and Survival Regression, with RPPA being a better indicator for the latter, while Gene Expression more closely aligned with PAM50. However, we believe that the results reflect the particular characteristics of our model and experimental setting, and should not be extrapolated without further validation.

Overall, our study has achieved \begin{enumerate*}[label=(\roman*)]
    \item a robust evaluation of the effect of dropout in a non-linear BINN; 
    \item integrated interpretability, with pathway activity scores enabling downstream analyses such as feature importance, clustering, and survival prediction;
    \item a fully interpretable classification case study;
    and \item a thorough multi-omics extension, revealing the positive impact of multi-omics integration, the impact of individual omics layers, and establishing practical guidelines for integration within our framework.
\end{enumerate*}.

As is common in complex modelling studies, our approach comes with certain limitations. We note the non-convex nature of deep neural network optimisation, which is a well-recognised limitation in the field and an area of on-going research. Although we have thoroughly studied the repeatability of our model and identified both artifact-specific and globally-significant pathways, it is important to acknowledge that deep learning models trained on relatively small datasets cannot be expected to predict the full complexity of biological relationships underlying a disease. Our study on latent space dropout, even though echoing results from previous work in terms of model repeatability \citep{fortelny_knowledge-primed_2020,seninge_vega_2021,selby_visible_2024}, also highlights limitations of such regularisation techniques. This observation opens up new avenues for investigating more principled approaches to improving model robustness and biological fidelity. Another promising direction for future work is the training of omics foundation models (e.g.\citep{yang_scbert_2022,chen_transformer_2023,cui_scgpt_2023,liu_transgem_2024}), which leverage larger datasets to enhance model stability and generalisability. However, these  models are mostly trained on single-cell data, and their applicability to bulk modalities remains an open question. Similarly, tabular foundation models (e.g.TabPFN \citep{hollmann_tabpfn_2023,hollmann_accurate_2025}) mark recent progress, but their deployment in high-dimensional omics context may face scalability challenges due to increased memory requirements and the comparatively lower dropout rates characteristic of bulk data.

\section*{Acknowledgements} We thank Roman Laddach for help with data acquisition, and Roman Laddach and Wai Yee Wong for fruitful discussions. MW and ST acknowledge funding from King’s College London and the A*STAR Research Attachment Programme (ARAP) to PHdCA. MW acknowledges funding by the AI, Analytics and Informatics (AI3) Horizontal Technology Programme Office (HTPO) seed grant (grant no: C211118015) from A*STAR, Singapore. ST acknowledges funding from the British Skin Foundation (006/R/22) and the UK Royal Society (IES\textbackslash R2\textbackslash 222084).

\bibliographystyle{plainnat} 
\bibliography{references_zotero}

\clearpage

\appendix

\onecolumn

\section*{Supplementary Material for Incorporating Prior Knowledge in Deep Learning Models via Pathway Activity Autoencoders}

\section{Alternatives for Omics-Pathway Mappings} \label{app:sec:omics-mapping}

We briefly considered other alternative databases for non-gene-based omics layers. While the dataset provided by \citet{wissel_hierarchical_2022} uses a ``merged'' Methylation layer, we considered whether to use Illumina's 27K or 450 CpG-to-gene mapping. Since all annotations in 27k were also present in the 450 annotation, we decided to proceed with the more up-to-date annotation file, especially since the 450 there are multiple genes for the same CpG site.

For $\mu$RNA, we briefly considered using \textit{predicted} target genes instead of the experimentally validated genes of miRTarBase2025 \texttt{miRTarBase\_SE\_WR} \citep{cui_mirtarbase_2025}. That is, we considered using information from either \texttt{miRDB} \citep{liu_prediction_2019,chen_mirdb_2020}, TargetScanHuman \citep{mcgeary_biochemical_2019}, or miRNet \citep{fan_mirnet_2016,fan_mirnetfunctional_2018,chang_mirnet_2020}. However, we noticed that when using predicted targets the number of target genes for a significant portion of $\mu$RNA measurements in our dataset grew significantly, and compounded over each gene possibly being into more than one pathway. This resulted in every $\mu$RNA being associated, on average, with 2.3 pathways when following \citet{chen_mirdb_2020}'s recommendations. This translated in the memory requirement for training an early integration model almost doubling when adding the $\mu$RNA to a model with only Gene Expression, CNV, Mutation, and RPPA, despite the number of unique features in the $\mu$RNA layer being less than 5\% of the total unique features.

\section{Autoencoder Architecture and Training Environment For RNA-Seq Experiments} \label{app:sec:arch-and-env}

We train the AE, VAE, PAAE, and PAVAE models using batch training for 1024 epochs with a learning rate of $10^{-4}$ using the Adam optimiser under the \texttt{pytorch} v1.12.0 and the \texttt{skorch} v0.11.0 frameworks, all other components of our pipeline are built using \texttt{scikit-learn} v1.0.2. These models are trained with a dropout rate of 50\%.

These models used rectified linear units as nonlinearities after each hidden layer except the last layer in each module, i.e., the last layer in the pathway encoders, and the latent space encoder and decoder are all simple linear layers without any nonlinearity. We do this to allow our model to output any real value, and, due to our network initialization method, to produce roughly normal outputs.

For the Variational models we use the loss scheduling/warm-up procedure described in the main paper, using $T_{s}=32$ and with annealing duration $128$ for $T_{e} = T_{s} + 128$ in our annealed scheduling function $S_{T_{s},T_{e}}(t)$. We use a gaussian distribution defined by $\mu$ and $\sigma$, from which the values of $z$ are sampled using the reparametrization trick to maintain differentiability \citep{kingma_auto-encoding_2014}.

For our baselines, we either used the default values when it was available (PASL) or performed grid search within a certain budget. For PACL we trained both with the original sizes, adapting the last layer to have a dimensionality of 64, as our models' latent space also had 64, while also adding layer sizes that mirrored our choices for the AE model. Since we could not fit the entire PACL model in memory given our setup, we limited the batch size to 64. For PathME we used a hidden size that mirrored our choice for the PAAE model's pathway activity sizes, while performing grid search on a smaller subset of the hyperparameter grid proposed by the authors. We had to perform a search on a smaller space since we performed a more exhaustive grid search instead of employing bayesian optimisation, which allowed us to explore more combinations (54) than in the original paper (50), while also removing redundant searches (such as searching with a dropout of 100\%). Due to a software incompatibility with the CUDA libraries in PathME's library, we had to run it using the CPU, and thus we limited the batch-size to the highest batch size tested in their model (32).

For all pipelines, we first perform an internal 8-fold cross validation for hyperparameter tuning, selecting the model with highest ROC AUC from a grid search in the parameters shown in Tab.~\ref{tab:gridsearch}, where the classifier was applied to the hidden representation vector (i.e., $z$ or $\mu$), which resulted in the following hyperparameters being chosen:

\begin{itemize}
    \item Autoencoder: Encoder with 2 layers with 128, and 64 units; using a Logistic Regression classifier.
    \item Pathway Activity Autoencoders: Pathway Activity Encoders with 2 layers with a 32-unit hidden layer; an Encoder with only a single 64-unit layer; and using a Logistic Regression classifier.
    \item Variational Autoencoder: Encoder with 2 layers with 128, and $2\times64$ units (64 for $\mu$ and 64 for $\sigma$); using a Logistic Regression classifier.
    \item Pathway Activity Variational Autoencoders: A linear Pathway Activity Encoder with only a single linear layer; Encoder with 2 layers with 128, and $2\times64$ units (64 for $\mu$ and 64 for $\sigma$); and using a Support Vector Machine classifier.
    \item PACL with $k=1$, and 1024 epochs. For the Hallmark Genes pathway set the best model had two hidden layers with 128 and 64 units and a Logistic Regression classifier was used. For the KEGG pathway set the best model had three hidden layers with 500, 200 and 64 units and a Support Vector Machine classifier was used.
    \item PathME with 16 hidden units, $\alpha=1$, $\lambda=10^{-9}$ and $0.001$ with a Logistic Regression Classifier.
\end{itemize}

We train the Logistic Regression models with L2 regularization with $C=1$, for a maximum of $100$ iterations using the lbfgs solver, and using the softmax function for the multi-class cases. For the Support Vector Machine classifiers we use $C=1$ as the l2 regularization parameter and a Radial Basis Function kernel. For the Random Forest Classifier we build 100 Trees using Gini impurity, considering $\sqrt{d_{input}}$ features for splitting, boostrapping samples when building trees.

\begin{sidewaystable}[bpht]
    \centering
    \footnotesize
    \begin{tabular}{cccccc}
         \toprule
         Model & Encoder Layer Sizes & $\beta$ & $\beta$ Schedule & Pathway Module Hidden Layer Sizes & Classifier\\
         \midrule
         AE & [128,64], [256,128,64], [512,256,128,64] & N/A & N/A & N/A & LR,SVM,RF \\
         VAE & [128,64], [256,128,64], [512,256,128,64] & 1,5,10,50,100 & 
         $\chi_{t>32}(t)$, $S_{32,160}(t)$ & N/A & LR,SVM,RF \\
         PAAE & [64],[128,64] & N/A & N/A & [],[32],[32,16] & LR,SVM,RF \\
         PAVAE & [128,64], [256,128,64] & 1,5,10,50,100 & $\chi_{t>32}(t)$, $S_{32,160}(t)$ & [],[32],[32,16] & LR,SVM,RF \\
        \midrule
         & Layer Sizes & Epochs & RBM k  & & \\
        \midrule
        PACL & [64], [128,64], [256,128,64], [500,200,64] & 16,64,1024,256 & 16,4,1 & & LR,SVM,RF \\
        \midrule
        & Layer Sizes & $\alpha$ & $\lambda$ & Learning Rate & \\
        \midrule
        PathME & [16],[32],[64] & 0,1 & $10^{-9},10^{-5},10^{-1}$ & $10^{-5},10^{-5},10^{-1}$ & LR,SVM,RF \\
        \midrule
        & a1+a2 & t &$\lambda$ & m & \\
        \midrule
        PASL & 350+150 & 0.9 & $\frac{1}{3}$ & 2000 & LR,SVM,RF \\
        \bottomrule
    \end{tabular}
    \caption{Hyperparameters used during the internal validation grid search.}
    \label{tab:gridsearch}
\end{sidewaystable}

After the hyperparameters were selected, we train the models in a completely unsupervised fashion on our training dataset (TCGA) and then use the models to compress the input to their latent representation, using $z$ for the AE and PAAE models and the posterior means $\mu$ for the VAE and PAVAE models, as well as a separate compression of our input to its pathway activity score representation $a$ for Pathway Activity models, then using this compressed latent representation as the input for training the classifier.

After training, for the AE, VAE, PAAE, and PAVAE models, we then normalize the input of our test dataset using the same method as was used for the training dataset, since our test dataset is on a completely different scale and was produced by a different gene expresssion measuring technology. We then use this as input for the all models to compress the features, and use the compressed representation as input for the classifiers that produce predictions of each sample's cancer subtypes.

We adapted the original code provided by PASL and PathME to run their models, and use a reimplementation of PACL to allow us to use the probability values instead of binary vectors as the internal representations, which allowed for better performance than the original code. It is of note that for PASL we used standard normalisation (z-scoring) since that is what was used internally in the model, and when it came to applying the model to an external dataset we z-scored the dataset and then multiplied it with the inverted model matrix, instead of using the models' learned mean and average. For PACL, since RBMs need a probability input, we z-scored the values as described in the original paper, but then performed 0-1 scaling so that each value could be treated as a probability for the training sampler. We could not run the PathME model on our GPU since the library it was developed on was incompatible with out setup. Furthermore, PathME had stability issues and produced \texttt{nan} or \texttt{inf} values for some pathways, whenever this happens we drop these pathways during the internal and external cross validation.

For all tests we consider the macro-averaged Area Under the Receiver Operating Characteristic Curve (ROC AUC) as the main metric and, unless otherwise stated, we apply the Wilcoxon Rank-Sum test for testing statistical significance between different medians. We also provide other metrics such as accuracy, precision, recall and F1 score. Note that our Pathway Activity methods can't access the full set of input genes when building their internal representation, while at the same time having to rebuild the entirety of the input.

\subsection{Model sizes and alternatives} \label{app:sec:model-sizes}

Although nominally some of our baselines seem to have smaller parameter counts than our models, that is mostly due to the fact that these models either (1) only reconstruct the input partially, which is what is done by PathME, or (1) don't reconstruct the input, i.e. PACL and PASL. One can see in Tables~\ref{tab:model-sizes-naive},~\ref{tab:model-sizes-hg}~and~\ref{tab:model-sizes-kegg} the model parameter sizes for every hyperparameter configuration. In Tables~\ref{tab:model-sizes-hg}~and~\ref{tab:model-sizes-kegg}, we show a comparison with other models when we consider \begin{enumerate*}[label={(\alph*)}]
    \item\label{lst:sz-cmp-full} that our PAAE model and PathME have to reconstruct the full input; then 
    \item\label{lst:sz-cmp-pway}  when we consider that our model only has to reconstruct the pathway set's member genes; and, finally,
    \item\label{lst:sz-cmp-enc} if we are only interested in dimensionality reduction, what would be our model size if we only kept the Encoding part and did not keep the reconstruction part.
\end{enumerate*}

For brevity's sake, we discuss here only the results for Table~\ref{tab:model-sizes-hg}, since those in Table~\ref{tab:model-sizes-kegg} are equivalent. One can see that if we were to reconstruct the full gene matrix using only pathway information \ref{lst:sz-cmp-full}, our PAAE model is vastly more efficient than using a PathME model, since the PathME model builds one separate autoencoder per pathway. If we were to use our PAAE model to only reconstruct genes that are contained within the pathway set \ref{lst:sz-cmp-pway}, then our PAAE model continues to be competitive w.r.t. to PathME, and we can still keep the performance from reconstructing the full gene matrix \ref{lst:sz-cmp-full} as we'd only need to prune the decoder, only needing these extra parameters for training. If we only need the encoder \ref{lst:sz-cmp-enc}, while our more complex models requires more parameters than PACL, they still vastly outperform it. Furthermore, our model requires less parameters than PASL, even though PASL's parameter count is pruned to only include nonzero values. Again, our model achieves this while still being able to keep the performance from reconstructing the full gene matrix \ref{lst:sz-cmp-full} since we'd only need to prune our model and discard these parameters after training.

\subsection{PACL and PASL external validation discussion} \label{app:sec:pacl-pasl-ext-val-discussion}

Both PACL and PASL experienced catastrophic failures when translating to another dataset. While PACL's failure on its latent space was expected due to the information loss caused by using locally-trained RBMs, its failure on the pathway space was unexpected. The same is also true for the PASL results. We posit that this is due to the normalisation methods used by these methods. Since PACL used z-scoring and since we also needed to 0-1 normalise the input, we chose to keep z-scoring and min-max scaling as its default scaling method. PASL similarly uses z-scoring, which we attempted to use natively using its own codebase, but which caused numerical errors and made the model fail altogether. We then proceeded to apply the model in Python instead of Matlab, which avoided the numerical errors but still produced subpar results. We left a comparison of improved versions of these methods' pipelines to future work due to computational budget constraints.

\begin{table}[]
    \centering
    \footnotesize
    \begin{tabular}{ccrcr}
        \toprule
        Model & Configuration & Parameters & Space & Dimensionality \\
        \midrule
        \multicolumn{5}{c}{No Prior Knowledge} \\
        \midrule
        \multirow{3}{*}{AE} & $[\left|G\right|,128,64,128,\left|G\right|]$ & 4,795,105  & $z$ & 64 \\
        & $[\left|G\right|,256,128,64,128,256,\left|G\right|]$ & 9,620,961 & $z$ & 64 \\
        & $[\left|G\right|,512,256,128,64,128,256,512,\left|G\right|]$ & 19,403,745 & $z$ & 64 \\
        \bottomrule
    \end{tabular}
    \caption{Model parameter sizes for every pararameter configuration tested in our hyperparameter search for the Naive (no prior knowledge) models. Please see Tables~\ref{tab:model-sizes-hg}~and~\ref{tab:model-sizes-kegg} for other models and Section~\ref{app:sec:model-sizes} for discussion on these results.}
    \label{tab:model-sizes-naive}
\end{table}

\begin{table}[]
    \centering
    \tiny
    \begin{tabular}{ccrcr}
        \toprule
        Model & Configuration & Parameters & Space & Dimensionality \\
        \midrule
        \multicolumn{5}{c}{Reconstruct Full Gene Matrix} \\ \midrule
        \multirow{12}{*}{PAAE} & \multirow{2}{*}{$\left( \left|P\right|\times[d_{p},1] \right) \rightarrow [64,\left|G\right|]$} & \multirow{2}{*}{1,219,041} & $z$ & 64 \\
         &&& $a$ & $\left|P\right|=50$ \\
         & \multirow{2}{*}{$\left( \left|P\right|\times[d_{p},32,1] \right) \rightarrow [64,\left|G\right|]$} & \multirow{2}{*}{1,444,883} & $z$ & 64 \\
         &&& $a$ & $\left|P\right|=50$ \\
         & \multirow{2}{*}{$\left( \left|P\right|\times[d_{p},32,16,1] \right) \rightarrow [64,\left|G\right|]$} & \multirow{2}{*}{1,470,483} & $z$ & 64 \\
         &&& $a$ & $\left|P\right|=50$ \\
         & \multirow{2}{*}{$\left( \left|P\right|\times[d_{p},1] \right) \rightarrow [128,64,128,\left|G\right|]$} & \multirow{2}{*}{2,428,833} & $z$ & 64 \\
         &&& $a$ & $\left|P\right|=50$ \\
         & \multirow{2}{*}{$\left( \left|P\right|\times[d_{p},32,1] \right) \rightarrow [128,64,128,\left|G\right|]$} & \multirow{2}{*}{2,654,675} & $z$ & 64 \\
         &&& $a$ & $\left|P\right|=50$ \\
         & \multirow{2}{*}{$\left( \left|P\right|\times[d_{p},32,16,1] \right) \rightarrow [128,64,128,\left|G\right|]$} & \multirow{2}{*}{2,680,275} & $z$ & 64 \\
         &&& $a$ & $\left|P\right|=50$ \\
    
        \multirow{5}{*}{PathME} & $\left|P\right|\times[d_{p},16,1,16,G]$ & 43,514,144 & $a$ & \multirow{4}{*}{$\left|P\right|=50$} \\
         & $\left|P\right|\times[d_{p},32,1,32,G]$ & 87,028,288 & $a$ &  \\
         & $\left|P\right|\times[d_{p},64,1,64,G]$ & 174,056,576 & $a$ &  \\
         & $\left|P\right|\times[d_{p},128,1,128,G]$ & 348,113,152 & $a$ &  \\
         & $\left|P\right|\times[d_{p},256,1,256,G]$ & 696,226,304 & $a$ &  \\
         
        \midrule
        \multicolumn{5}{c}{Reconstructing Only Pathway's Input Genes} \\ \midrule
         
        \multirow{12}{*}{PAAE} & \multirow{2}{*}{$\left( \left|P\right|\times[d_{p},1] \right) \rightarrow [64,\left|G\right|]$} & \multirow{2}{*}{289,411} & $z$ & 64 \\
         &&& $a$ & $\left|P\right|=50$ \\
         & \multirow{2}{*}{$\left( \left|P\right|\times[d_{p},32,1] \right) \rightarrow [64,\left|G\right|]$} & \multirow{2}{*}{515,253} & $z$ & 64 \\
         &&& $a$ & $\left|P\right|=50$ \\
         & \multirow{2}{*}{$\left( \left|P\right|\times[d_{p},32,16,1] \right) \rightarrow [64,\left|G\right|]$} & \multirow{2}{*}{540,853} & $z$ & 64 \\
         &&& $a$ & $\left|P\right|=50$ \\
         & \multirow{2}{*}{$\left( \left|P\right|\times[d_{p},1] \right) \rightarrow [128,64,128,\left|G\right|]$} & \multirow{2}{*}{583,875} & $z$ & 64 \\
         &&& $a$ & $\left|P\right|=50$ \\
         & \multirow{2}{*}{$\left( \left|P\right|\times[d_{p},32,1] \right) \rightarrow [128,64,128,\left|G\right|]$} & \multirow{2}{*}{809,717} & $z$ & 64 \\
         &&& $a$ & $\left|P\right|=50$ \\
         & \multirow{2}{*}{$\left( \left|P\right|\times[d_{p},32,16,1] \right) \rightarrow [128,64,128,\left|G\right|]$} & \multirow{2}{*}{835,317} & $z$ & 64 \\
         &&& $a$ & $\left|P\right|=50$ \\
    
        \multirow{5}{*}{PathME} & $\left|P\right|\times[d_{p},16,1,16,d_{p}]$ & 233,088 & $a$ & \multirow{4}{*}{$\left|P\right|=50$} \\
         & $\left|P\right|\times[d_{p},32,1,32,d_{p}]$ & 466,176 & $a$ &  \\
         & $\left|P\right|\times[d_{p},64,1,64,d_{p}]$ & 932,352 & $a$ &  \\
         & $\left|P\right|\times[d_{p},128,1,128,d_{p}]$ & 1,864,704 & $a$ &  \\
         & $\left|P\right|\times[d_{p},256,1,256,d_{p}]$ & 3,729,408 & $a$ &  \\
         
        \midrule
        \multicolumn{5}{c}{Encoder Only} \\ \midrule
        \multirow{12}{*}{PAAE} & \multirow{2}{*}{$\left( \left|P\right|\times[d_{p},1] \right) \rightarrow [64,\left|G\right|]$} & 10,496 & $z$ & 64 \\
         && 7,232 & $a$ & $\left|P\right|=50$ \\
         & \multirow{2}{*}{$\left( \left|P\right|\times[d_{p},32,1] \right) \rightarrow [64,\left|G\right|]$} & 236,338 & $z$ & 64 \\
         && 233,074 & $a$ & $\left|P\right|=50$ \\
         & \multirow{2}{*}{$\left( \left|P\right|\times[d_{p},32,16,1] \right) \rightarrow [64,\left|G\right|]$} & 261,938 & $z$ & 64 \\
         && 258,674 & $a$ & $\left|P\right|=50$ \\
         & \multirow{2}{*}{$\left( \left|P\right|\times[d_{p},1] \right) \rightarrow [128,64,128,\left|G\right|]$} & 22,016 & $z$ & 64 \\
         && 7,232 & $a$ & $\left|P\right|=50$ \\
         & \multirow{2}{*}{$\left( \left|P\right|\times[d_{p},32,1] \right) \rightarrow [128,64,128,\left|G\right|]$} & 247,858 & $z$ & 64 \\
         && 233,074 & $a$ & $\left|P\right|=50$ \\
         & \multirow{2}{*}{$\left( \left|P\right|\times[d_{p},32,16,1] \right) \rightarrow [128,64,128,\left|G\right|]$} & 273,458 & $z$ & 64 \\
         && 258,674 & $a$ & $\left|P\right|=50$ \\
    
        \multirow{8}{*}{PACL} & \multirow{2}{*}{$\left( \left|P\right|\times[d_{p},1] \right) \rightarrow [64]$} & 14,837 & $z$ & 64 \\
         && 11,523 & $a$ & $\left|P\right|=50$ \\
         & \multirow{2}{*}{$\left( \left|P\right|\times[d_{p},1] \right) \rightarrow [128,64]$} & 26,485 & $z$ & 64 \\
         && 11,523 & $a$ & $\left|P\right|=50$ \\
         & \multirow{2}{*}{$\left( \left|P\right|\times[d_{p},1] \right) \rightarrow [256,128,64]$} & 66,165 & $z$ & 64 \\
         && 11,523 & $a$ & $\left|P\right|=50$ \\
         & \multirow{2}{*}{$\left( \left|P\right|\times[d_{p},1] \right) \rightarrow [500,200,64]$} & 150,837 & $z$ & 64 \\
         && 11,523 & $a$ & $\left|P\right|=50$ \\
        
         \multirow{3}{*}{PASL} & \multirow{3}{*}{$[D]=[D_{1}+D_{2}]=[500]=[350+150]$} & \multirow{3}{*}{354,197} & $a'$ & $D_{1}=350$ \\
         &&& $z'$ & $D_{2}=150$ \\
         &&& $L$ & $D=500$ \\
        \bottomrule
    \end{tabular}
    \caption{Model parameter sizes for every pararameter configuration tested in our hyperparameter search for the models using the Hallmark Genes pathway set. Please see Tables~\ref{tab:model-sizes-naive}~and~\ref{tab:model-sizes-kegg} for other models and Section~\ref{app:sec:model-sizes} for discussion on these results.}
    \label{tab:model-sizes-hg}
\end{table}

\begin{table}[]
    \centering
    \tiny
    \begin{tabular}{ccrcr}
        \toprule
        Model & Configuration & Parameters & Space & Dimensionality \\
        \midrule
        \multicolumn{5}{c}{Reconstruct Full Gene Matrix} \\ \midrule
        \multirow{12}{*}{PAAE} & \multirow{2}{*}{$\left( \left|P\right|\times[d_{p},1] \right) \rightarrow [64,\left|G\right|]$} & \multirow{2}{*}{1,233,085} & $z$ & 64 \\
         &  &  & $a$ & $\left|P\right|=186$ \\
         & \multirow{2}{*}{$\left( \left|P\right|\times[d_{p},32,1] \right) \rightarrow [64,\left|G\right|]$} & \multirow{2}{*}{1,628,955} & $z$ & 64 \\
         &  &  & $a$ & $\left|P\right|=186$ \\
         & \multirow{2}{*}{$\left( \left|P\right|\times[d_{p},32,16,1] \right) \rightarrow [64,\left|G\right|]$} & \multirow{2}{*}{1,724,187} & $z$ & 64 \\
         &  &  & $a$ & $\left|P\right|=186$ \\
         & \multirow{2}{*}{$\left( \left|P\right|\times[d_{p},1] \right) \rightarrow [128,64,128,\left|G\right|]$} & \multirow{2}{*}{2,451,581} & $z$ & 64 \\
         &  &  & $a$ & $\left|P\right|=186$ \\
         & \multirow{2}{*}{$\left( \left|P\right|\times[d_{p},32,1] \right) \rightarrow [128,64,128,\left|G\right|]$} & \multirow{2}{*}{2,847,451} & $z$ & 64 \\
         &  &  & $a$ & $\left|P\right|=186$ \\
         & \multirow{2}{*}{$\left( \left|P\right|\times[d_{p},32,16,1] \right) \rightarrow [128,64,128,\left|G\right|]$} & \multirow{2}{*}{2,942,683} & $z$ & 64 \\
         &  &  & $a$ & $\left|P\right|=186$ \\
         
        \multirow{4}{*}{PathME} & $\left|P\right|\times[d_{p},16,1,16,G]$ & 161,643,344 & $a$ & \multirow{4}{*}{$\left|P\right|=186$} \\
         & $\left|P\right|\times[d_{p},32,1,32,G]$ & 323,286,688 & $a$ &  \\
         & $\left|P\right|\times[d_{p},64,1,64,G]$ & 646,573,376 & $a$ &  \\
         & $\left|P\right|\times[d_{p},128,1,128,G]$ & 1,293,146,752 & $a$ & \\
    
        \midrule
        \multicolumn{5}{c}{Reconstruct Only Pathway's Input Genes} \\ \midrule
        
        \multirow{12}{*}{PAAE} & \multirow{2}{*}{$\left( \left|P\right|\times[d_{p},1] \right) \rightarrow [64,\left|G\right|]$} & \multirow{2}{*}{350,385} & $z$ & 64 \\
         &  &  & $a$ & $\left|P\right|=186$ \\
         & \multirow{2}{*}{$\left( \left|P\right|\times[d_{p},32,1] \right) \rightarrow [64,\left|G\right|]$} & \multirow{2}{*}{746,255} & $z$ & 64 \\
         &  &  & $a$ & $\left|P\right|=186$ \\
         & \multirow{2}{*}{$\left( \left|P\right|\times[d_{p},32,16,1] \right) \rightarrow [64,\left|G\right|]$} & \multirow{2}{*}{841,487} & $z$ & 64 \\
         &  &  & $a$ & $\left|P\right|=186$ \\
         & \multirow{2}{*}{$\left( \left|P\right|\times[d_{p},1] \right) \rightarrow [128,64,128,\left|G\right|]$} & \multirow{2}{*}{699,761} & $z$ & 64 \\
         &  &  & $a$ & $\left|P\right|=186$ \\
         & \multirow{2}{*}{$\left( \left|P\right|\times[d_{p},32,1] \right) \rightarrow [128,64,128,\left|G\right|]$} & \multirow{2}{*}{1,095,631} & $z$ & 64 \\
         &  &  & $a$ & $\left|P\right|=186$ \\
         & \multirow{2}{*}{$\left( \left|P\right|\times[d_{p},32,16,1] \right) \rightarrow [128,64,128,\left|G\right|]$} & \multirow{2}{*}{1,190,863} & $z$ & 64 \\
         &  &  & $a$ & $\left|P\right|=186$ \\
        
        \multirow{4}{*}{PathME} & $\left|P\right|\times[d_{p},16,1,16,d_{p}]$ & 408,544 & $a$ & \multirow{4}{*}{$\left|P\right|=186$} \\
         & $\left|P\right|\times[d_{p},32,1,32,d_{p}]$ & 817,088 & $a$ &  \\
         & $\left|P\right|\times[d_{p},64,1,64,d_{p}]$ & 1,634,176 & $a$ &  \\
         & $\left|P\right|\times[d_{p},128,1,128,d_{p}]$ & 3,268,352 & $a$ & \\
    
        \midrule
        \multicolumn{5}{c}{Encoder Only} \\ \midrule
        
        \multirow{12}{*}{PAAE} & \multirow{2}{*}{$\left( \left|P\right|\times[d_{p},1] \right) \rightarrow [64,\left|G\right|]$} & 24,540 & $z$ & 64 \\
         &  & 12,572 & $a$ & $\left|P\right|=186$ \\
         & \multirow{2}{*}{$\left( \left|P\right|\times[d_{p},32,1] \right) \rightarrow [64,\left|G\right|]$} & 420,410 & $z$ & 64 \\
         &  & 408,442 & $a$ & $\left|P\right|=186$ \\
         & \multirow{2}{*}{$\left( \left|P\right|\times[d_{p},32,16,1] \right) \rightarrow [64,\left|G\right|]$} & 515,642 & $z$ & 64 \\
         &  & 503,674 & $a$ & $\left|P\right|=186$ \\
         & \multirow{2}{*}{$\left( \left|P\right|\times[d_{p},1] \right) \rightarrow [128,64,128,\left|G\right|]$} & 44,764 & $z$ & 64 \\
         &  & 12,572 & $a$ & $\left|P\right|=186$ \\
         & \multirow{2}{*}{$\left( \left|P\right|\times[d_{p},32,1] \right) \rightarrow [128,64,128,\left|G\right|]$} & 440,634 & $z$ & 64 \\
         &  & 408,442 & $a$ & $\left|P\right|=186$ \\
         & \multirow{2}{*}{$\left( \left|P\right|\times[d_{p},32,16,1] \right) \rightarrow [128,64,128,\left|G\right|]$} & 535,866 & $z$ & 64 \\
         &  & 503,674 & $a$ & $\left|P\right|=186$ \\
    
        \multirow{8}{*}{PACL} & \multirow{2}{*}{$\left( \left|P\right|\times[d_{p},1] \right) \rightarrow [64]$} & 29,739 & $z$ & 64 \\
         && 17,585 & $a$ & $\left|P\right|=186$ \\
         & \multirow{2}{*}{$\left( \left|P\right|\times[d_{p},1] \right) \rightarrow [128,64]$} & 50,091 & $z$ & 64 \\
         && 17,585 & $a$ & $\left|P\right|=186$ \\
         & \multirow{2}{*}{$\left( \left|P\right|\times[d_{p},1] \right) \rightarrow [256,128,64]$} & 107,179 & $z$ & 64 \\
         && 17,585 & $a$ & $\left|P\right|=186$ \\
         & \multirow{2}{*}{$\left( \left|P\right|\times[d_{p},1] \right) \rightarrow [500,200,64]$} & 225,035 & $z$ & 64 \\
         && 17,585 & $a$ & $\left|P\right|=186$ \\
         
         \multirow{3}{*}{PASL} & \multirow{3}{*}{$[D]=[D_{1}+D_{2}]=[500]=[350+150]$} & \multirow{3}{*}{330,038} & $a'$ & $D_{1}=350$ \\
         &&& $z'$ & $D_{2}=150$ \\
         &&& $L$ & $D=500$ \\
        \bottomrule
    \end{tabular}
    \caption{Model parameter sizes for every pararameter configuration tested in our hyperparameter search for the models using the KEGG pathway set. Please see Tables~\ref{tab:model-sizes-naive}~and~\ref{tab:model-sizes-hg} for other models and Section~\ref{app:sec:model-sizes} for discussion on these results.}
    \label{tab:model-sizes-kegg}
\end{table}

\section{Interpreting the Pathway Activity Scores} \label{app:sec:interpret-pa}

As we can see in the original PAAE description \citep{da_costa_avelar_pathway_2024}, the pathway activity space that we visualise is built of the individual pathway activity scores of each pathway. Each of the pathway activity scores is built by stacking fully connected layers using the normalised gene expression of the genes belonging to a specific pathway as an input. The encoder in $E_{p_{j}}$, if it has $k$ layers, can be described as in Eq.~\ref{eq:paae-activity-encoder}:

\begin{equation}\label{eq:paae-activity-encoder}
\begin{split}
    a_{j} = E_{p_{j}}(x) &= h_{j,k} \\
    h_{j,k} &= W_{j,k} \times h_{j,k-1} + b_{j,k} \\
    h_{j,i} &= f_{j,k}(W_{j,i} \times h_{j,i-1} + b_{j,i}) \\
    h_{j,1} &= f_{j,k}(W_{j,1} \times x_{:,p_{j}} + b_{j,1})
\end{split}
\end{equation}

Where $W_{j,i}$ is a $\Re^{d_{j,i-1} \times d_{j,i}}$ matrix, and $f_{j,k}$ is a nonlinear function, not present in the last layer so that our output is roughly normal-shaped.

Since the weights $W_{j,i}$ can be both positive and negative, one should not expect the final result $a_{j}$ to have a positive/negative directionality w.r.t. the pathway's input $x_{:,p_{j}}$, and one should always consider the pathway activity scores in terms of \textbf{contrasts} between samples. If there is a biologically known directionality, then that can be used to either confirm or align the directionality of the pathway activity score by inverting its sign.

The pathway activity score $a_{j}$ is the value that we use in both our clustermap (Figs.~\ref{fig:clustermap-kegg}, \ref{fig:clustermap-hallmark}) and featuremap (Fig.~\ref{fig:je-featuremap}) visualisations, and thus both these visualisations should be interpreted with the caveat above. Furthermore, in cases where nonlinear layers are involved in the calculation of the pathway activity score $a_{j}$, one should not expect the values to behave linearly.

\section{Clustermaps} \label{app:sec:clustermaps}

\begin{figure*}
    \centering
    \subcaptionbox{TCGA (train)\label{fig:clustermap-hallmark:sub:tcga}}{\includegraphics[width=.49\linewidth]{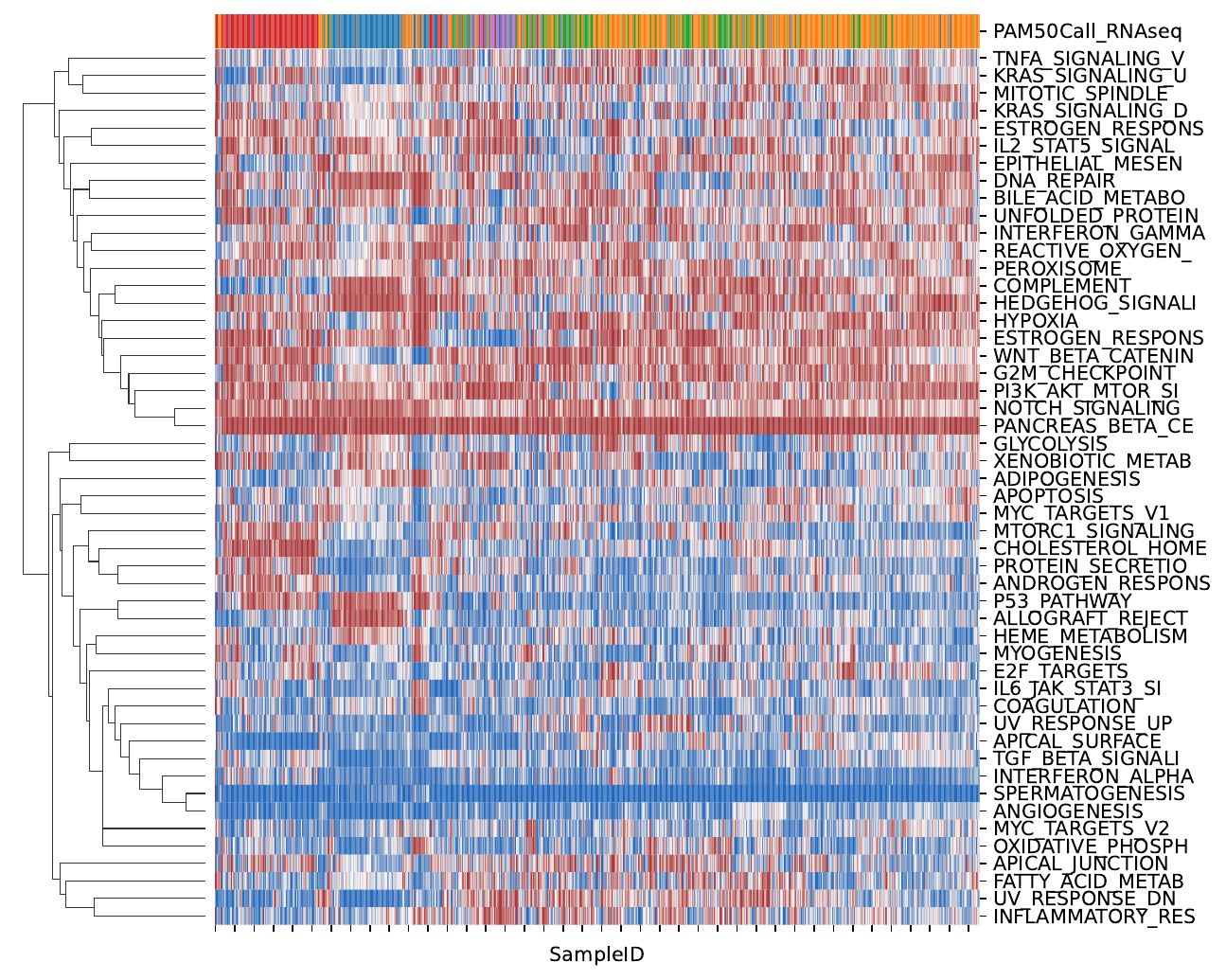}}
    \hfill
    \subcaptionbox{Metabric (test)\label{fig:clustermap-hallmark:sub:meta}}{\includegraphics[width=.49\linewidth]{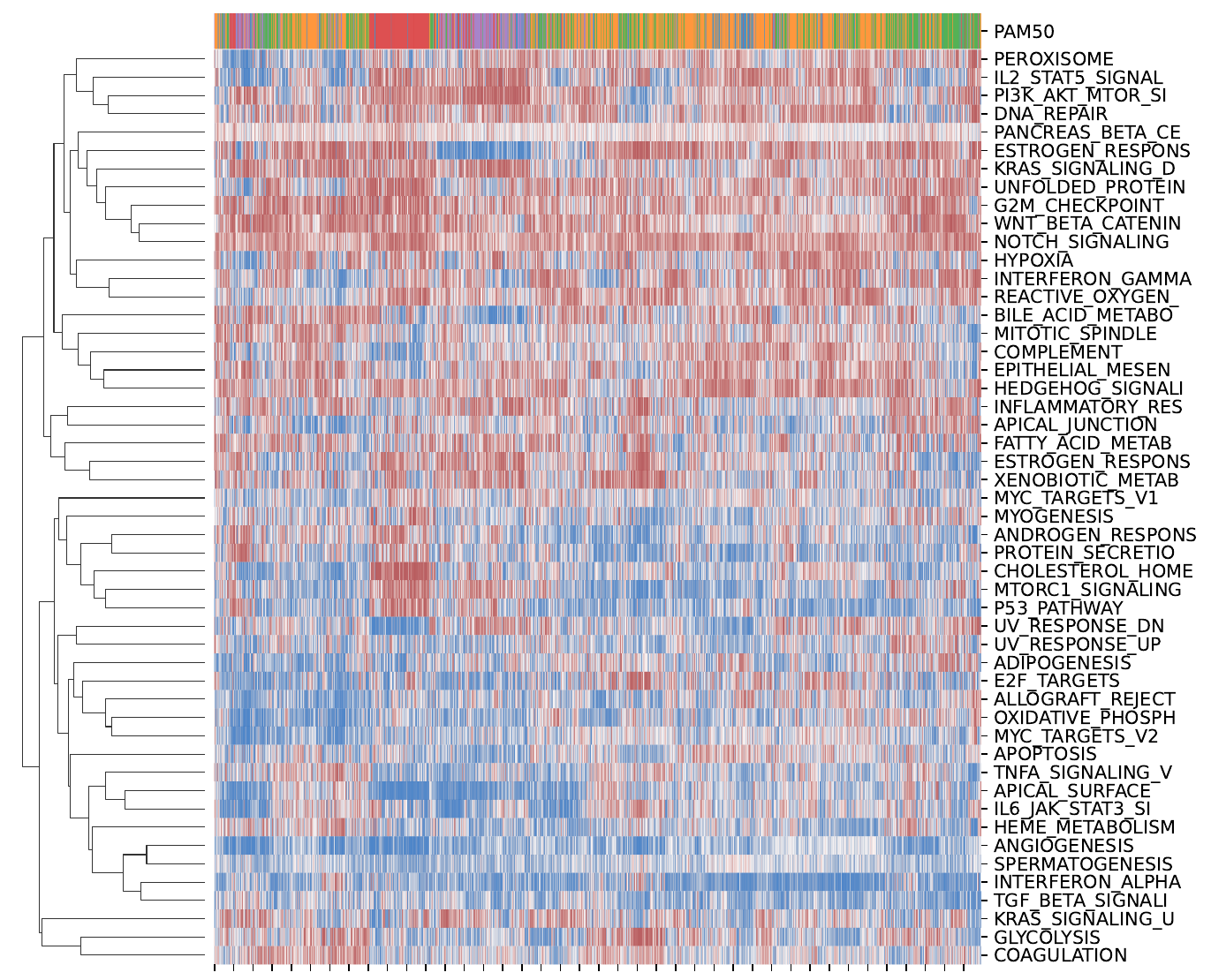}}
    
    \caption{The clustermap using the cosine distance between samples' inferred pathway activity vectors for Hallmark Genes PAAE's pathway activity space with the colours marking BRCA's 5-classes: {\color{brca-normal}Normal in blue}, {\color{brca-luma}Luminal A in orange}, {\color{brca-lumb}Luminal B in green}, {\color{brca-basal}Basal in red}, {\color{brca-her2}Her2 in purple}. Best seen in color.}
    \label{fig:clustermap-hallmark}
\end{figure*}

\begin{figure*}
    \centering
    \subcaptionbox{TCGA (train)\label{fig:clustermap-kegg:sub:tcga}}{\includegraphics[width=.45\linewidth]{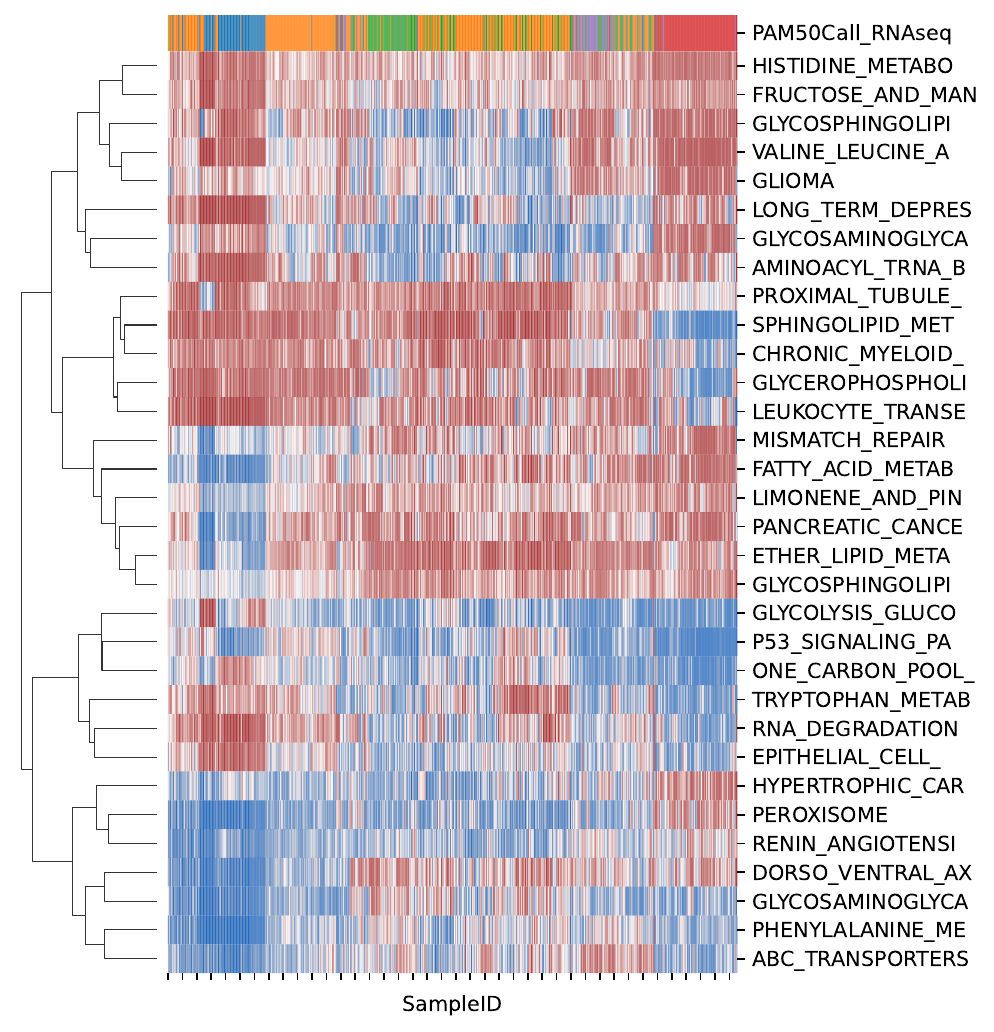}}
    \hfill
    \subcaptionbox{Metabric (test)\label{fig:clustermap-kegg:sub:meta}}{\includegraphics[width=.45\linewidth]{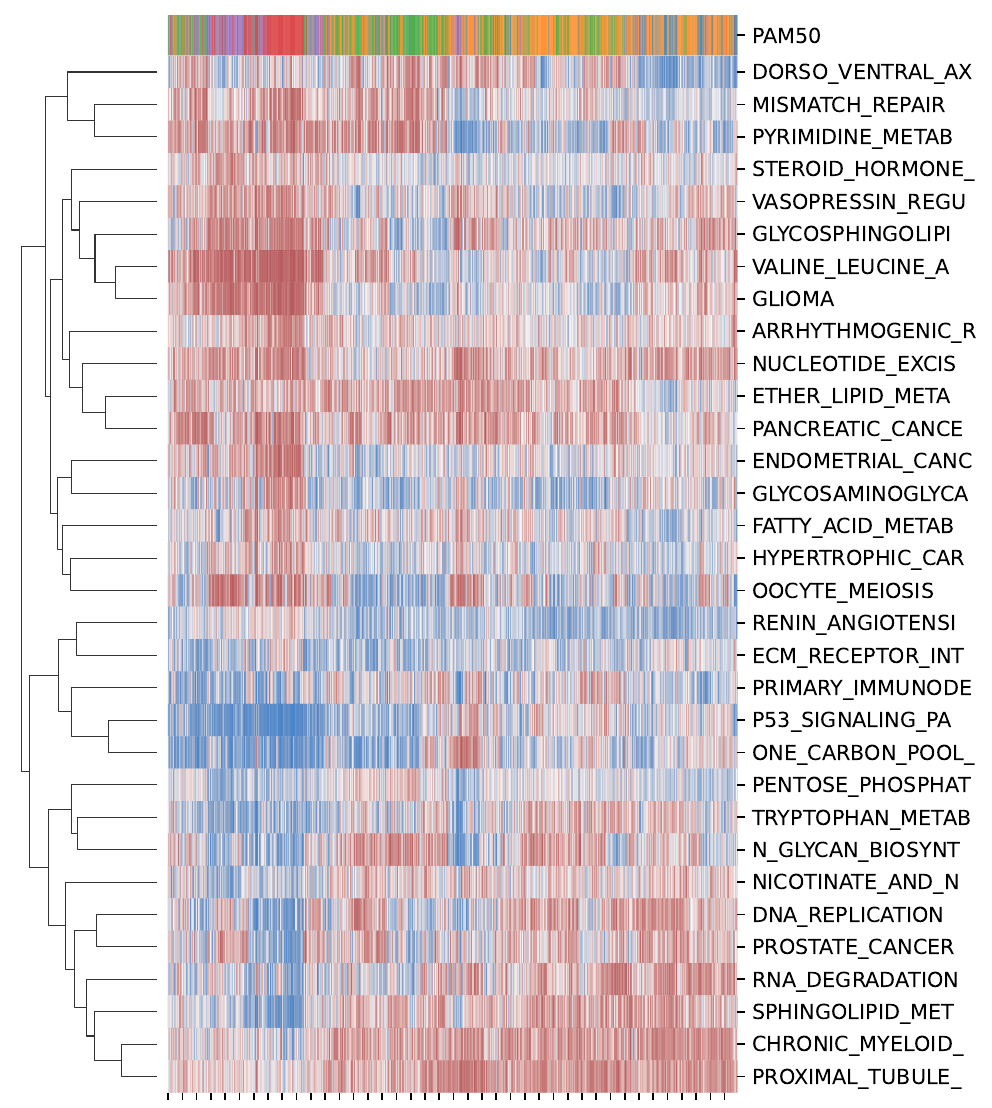}}
    
    \caption{The clustermap using the cosine distance between samples' inferred pathway activity vectors for KEGG PAAE's pathway activity space with the colours marking BRCA's 5-classes: {\color{brca-normal}Normal in blue}, {\color{brca-luma}Luminal A in orange}, {\color{brca-lumb}Luminal B in green}, {\color{brca-basal}Basal in red}, {\color{brca-her2}Her2 in purple}. Best seen in color.}
    \label{fig:clustermap-kegg}
\end{figure*}

For clustermaps, in Figs.~\ref{fig:clustermap-kegg} and \ref{fig:clustermap-hallmark}, we cluster along both rows and columns, using the cosine distance metric, as it is frequently used for deep learning embeddings, and we use \texttt{seaborn}'s \texttt{clustermap} function.

\section{Mutual Information} \label{app:sec:mutualinfo}

We calculate non-OvR Mutual Information using \texttt{scikit-learn}'s \texttt{mutual\_info\_classif} function, inside the \texttt{feature\_selection} module, to measure the mutual information between a continuous variable and the discrete cancer subtype labels. We do not filter for mutual information amongst variables, so multiple features might have a high MI w.r.t. to one class, instead of having multiple features with each having high MI w.r.t. to different classes. This differs from the OvR mutual information used in other parts of the paper.

\section{Survival Analyses} \label{app:sec:survival-analyses}

For survival plots and statistical tests (Fig.~\ref{fig:je-survival-kegg-importantgenes}) a time window limit of 5 years ($365\times5=1825$ days) was set. We use the \texttt{lifelines} library for both the tests and the plots. All of the analyses are done in an almost-unsupervised fashion, being that the only supervision we use is that we choose the top 5 pathways w.r.t. mutual information with the PAM50 subtypes and, since some subtypes might have different survival outcomes, this might influence the survival curves' relationship with survival.

For survival curves we select the most important genes from each pathway (as per ANPW, see \ref{app:sec:feat-importance}). For each of these genes, we divide the samples into ``high'' and ``low'' expression, which are defined as the upper and lower third quantiles of the TPM/IPM values (See ~\ref{app:sec:gene-norm} for details). We then perform logrank tests and select those genes that have a significant logrank p-value on the training dataset. Then we further filter to those genes which also have both a significant logrank p-value on the test dataset, and matching low-high survival sign.

For the low-high survival sign, we measure the surival rate at the end of the time window limit. If the ``high'' expression group has \textbf{higher} survival rate than the ``low'' expression group in the \textit{training} dataset, then it has a matching survival sign if it also has \textbf{higher} survival rate in the \textit{test} dataset. The same logic follows that if the ``high'' expression group has \textbf{lower} survival rate than the ``low'' expression group in the \textit{training} dataset, then it has a matching survival sign if it also has \textbf{lower} survival rate in the \textit{test} dataset.

\subsection{Preprocessing for Gene Expression Comparisons} \label{app:sec:gene-norm}

For the analyses involving raw input features, such as those in Fig.~\ref{fig:je-survival-kegg-importantgenes}, we re-normalise the genes from $\log(FPKM+1)$ to $\log(TPM+1)$ in the TCGA dataset, and from $\log(Intensity+1)$ to $\log(IPM+1)$, where FPKM stands for Fragments Per Kilobase Million, TPM for Transcripts Per Million, Intensity is the raw intensity value measured in the Metabric dataset, and IPM is ``Intensity Per Million'', where the intensity is normalised per sample as if it was a raw count value with $IPM_{g} = \frac{I_{g}\times10^{6}}{\sum_{g}I_{g}}$.

\subsection{Feature Importance} \label{app:sec:feat-importance}

Furthermore, we can get the most important features for each pathway using either neural path weights (NPW, Eq.~\ref{eq:npw}) or absolute neural path weight (ANPW, Eq.~\ref{eq:anpw}) \citep{uyar_multi-omics_2021}.

\begin{equation}\label{eq:npw}
    \operatorname{NPW}_{j} = \prod_{i=1}^{k} W_{j,i} ~,~ \operatorname{NPW}_{j} \in \Re^{|p_{j}|}
\end{equation}

\begin{equation}\label{eq:anpw}
    \operatorname{ANPW}_{j}[g] = |\operatorname{NPW}_{j}[g]|
\end{equation}

Note, however, that since our models might compute the pathway activity score value through a combination of nonlinear functions, these values should not be interpreted as a linear influence on the value, and combinations with other values may suppress or excite the score more than simply increasing/decreasing the genes with highest ANPW values.

Also note that one can do this interpretability technique for any point in any neural network, but we only define this for our pathway activity score for simplicity sake.

\section{Supporting Figures}

\begin{figure}
    \centering
    \includegraphics[width=0.45\linewidth]{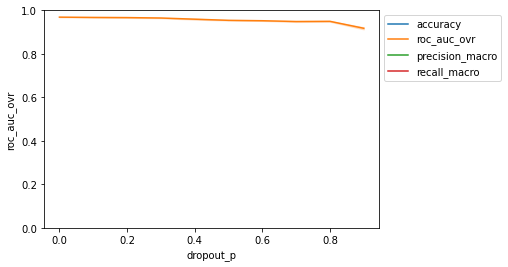}
    \includegraphics[width=0.45\linewidth]{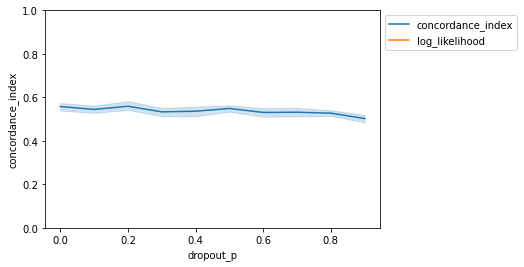}
    \caption{Average ROC AUC (left) from a one-vs-rest Logistic Regression and Concordance Index (right) of a Cox-PH model trained on the pathway activity vector of a PAAE model using the Hallmark Genes pathway set, trained on the \texttt{MO-TCGA-BRCA} dataset. Performance in shown across varying dropout rates applied on the pathway-activity layer. Both metrics show a gentle downward slope, with a sharper decrease between 70 and 80\% dropout. Shaded regions indicate model variance.}
    \label{fig:dropout-performance}
\end{figure}

\begin{figure}
    \centering
    \includegraphics[width=0.95\linewidth]{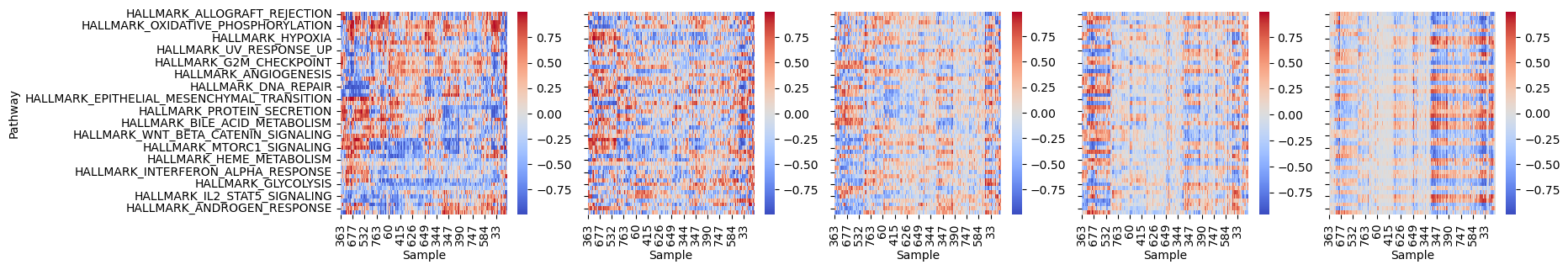} \\
    \includegraphics[width=0.95\linewidth]{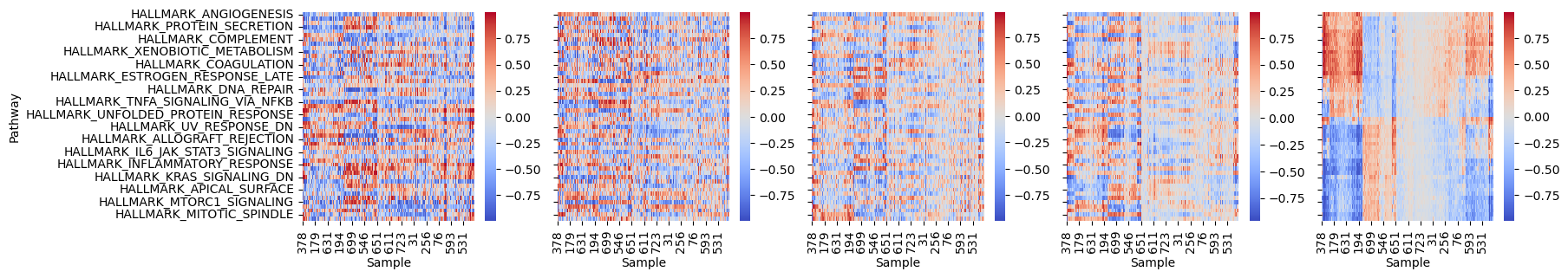}
    \caption{The learned pathway space when applying, from left to right, 0\%, 10\%, 50\%, 80\%, and 90\% dropout on the Pathway Activity Layer, clustered according to the result from 0\% (top 5) and 90\% dropout (bottom 5). One can see that, although dropout on the latent space has the ability to improve reproducibility, this comes at the cost of ``bleaching'' the learned representation.}
    \label{fig:dropout-heatmap-order}
\end{figure}

\begin{figure}
    \centering
    \includegraphics[width=0.95\linewidth]{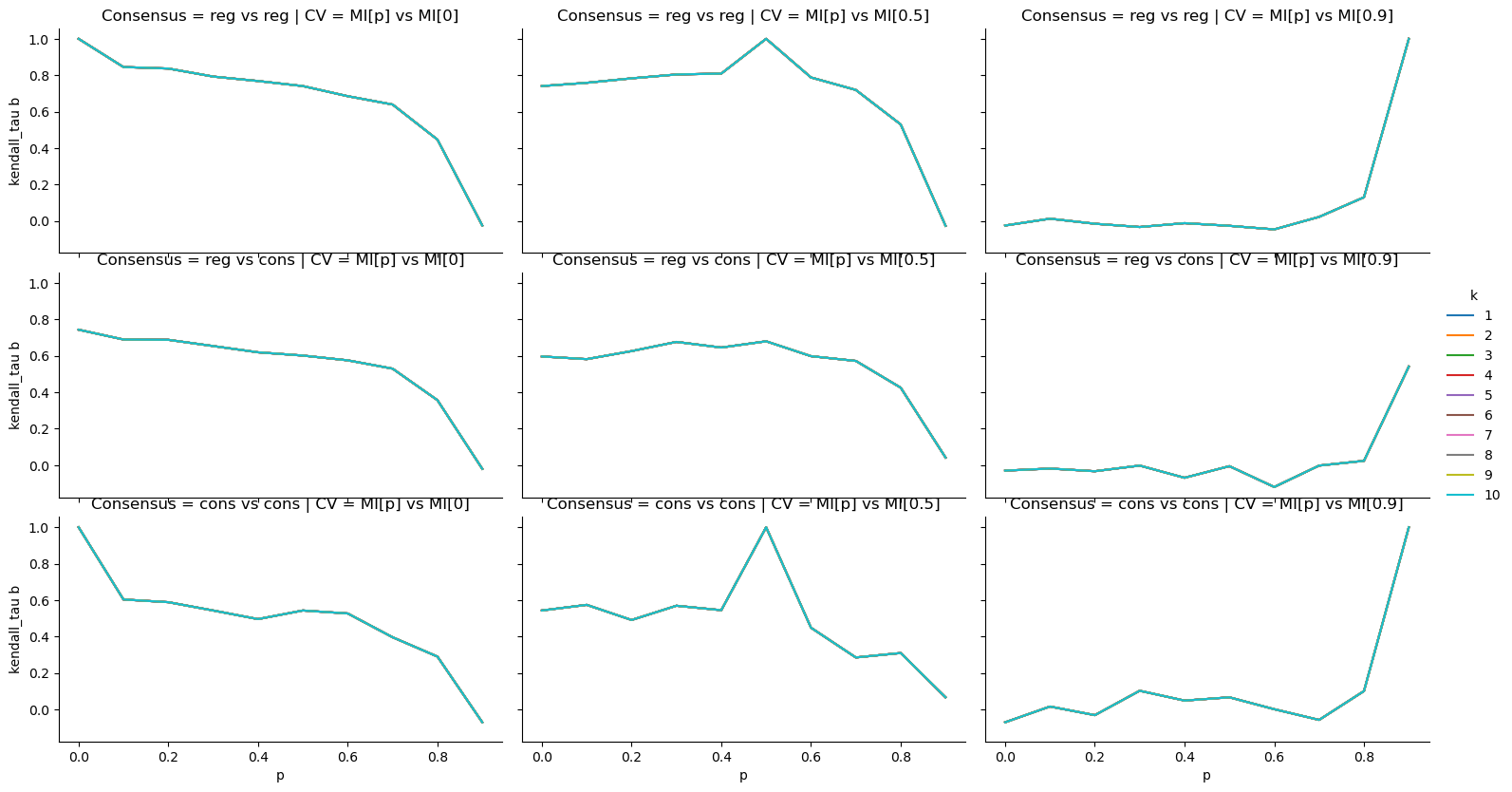}
    \caption{The average Kendall $\tau$-b coefficient between pathway rankings obtained from models trained with different random seeds on the \texttt{MO-TCGA-BRCA} dataset, based on mutual information. We compare non-consensus (\texttt{reg}) models between themselves (top 3), non-consensus against aligned consensus models (\texttt{cons}) and ACMs among themselves. The correlation is taken between models with $p$ dropout (x-axis) and models with $p=0$ (left), $p=0.5$ (middle) and $p=0.9$ (right) dropout. We can see that representational similarity seems to be consistent until a phase transition happens between 70\% and 80\% dropout.}
    \label{fig:dropout-kendall-tau-mi}
\end{figure}

\begin{figure}
    \centering
    \includegraphics[width=0.95\linewidth]{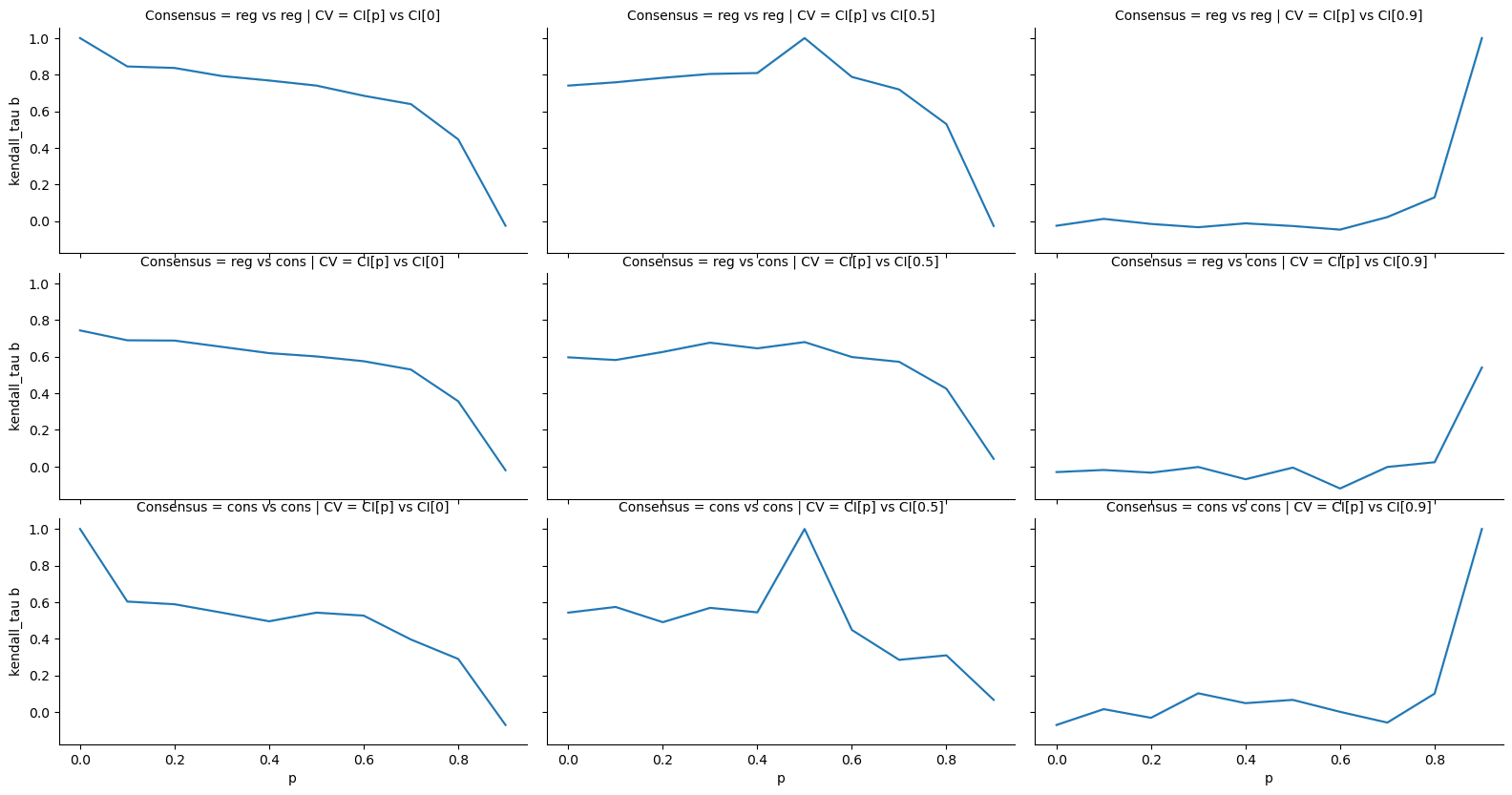}
    \caption{The average Kendall $\tau$-b coefficient between pathway rankings obtained from models trained with different random seeds on the \texttt{MO-TCGA-BRCA} dataset, based on concordance index. We compare non-consensus (\texttt{reg}) models between themselves (top 3), non-consensus against aligned consensus models (\texttt{cons}) and ACMs among themselves. The correlation is taken between models with $p$ dropout (x-axis) and models with $p=0$ (left), $p=0.5$ (middle) and $p=0.9$ (right) dropout. We can see that representational similarity seems to be consistent until a phase transition happens between 70\% and 80\% dropout.}
    \label{fig:dropout-kendall-tau-ci}
\end{figure}

\begin{figure}
    \centering
    \includegraphics[width=0.95\linewidth]{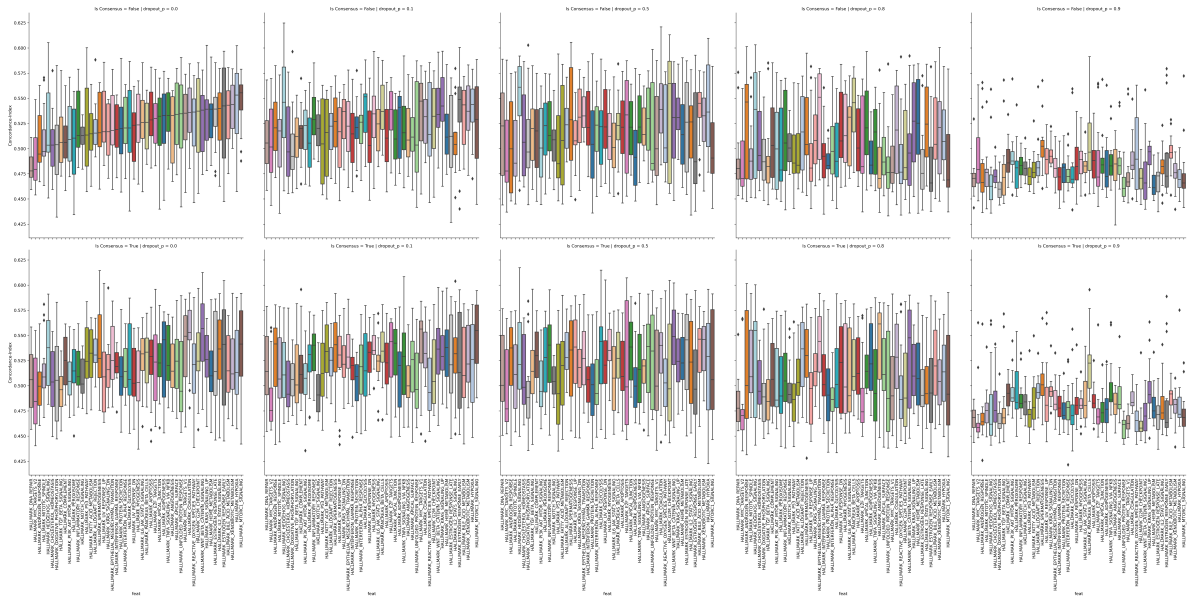}
    \caption{Boxplots of the concordance index of a pathway obtained from models trained with different random seeds on the \texttt{MO-TCGA-BRCA} dataset. Models are labelled as either Aligned Consensus Model (bottom) or not (top). From left to right, we show the results for dropout $p$ being $0.0$, $0.1$, $0.5$, $0.8$, and $0.9$. Note that model order changes drastically at the last two dropout values.}
    \label{fig:dropout-bothci}
\end{figure}

\begin{figure}
    \centering
    \includegraphics[width=0.95\linewidth]{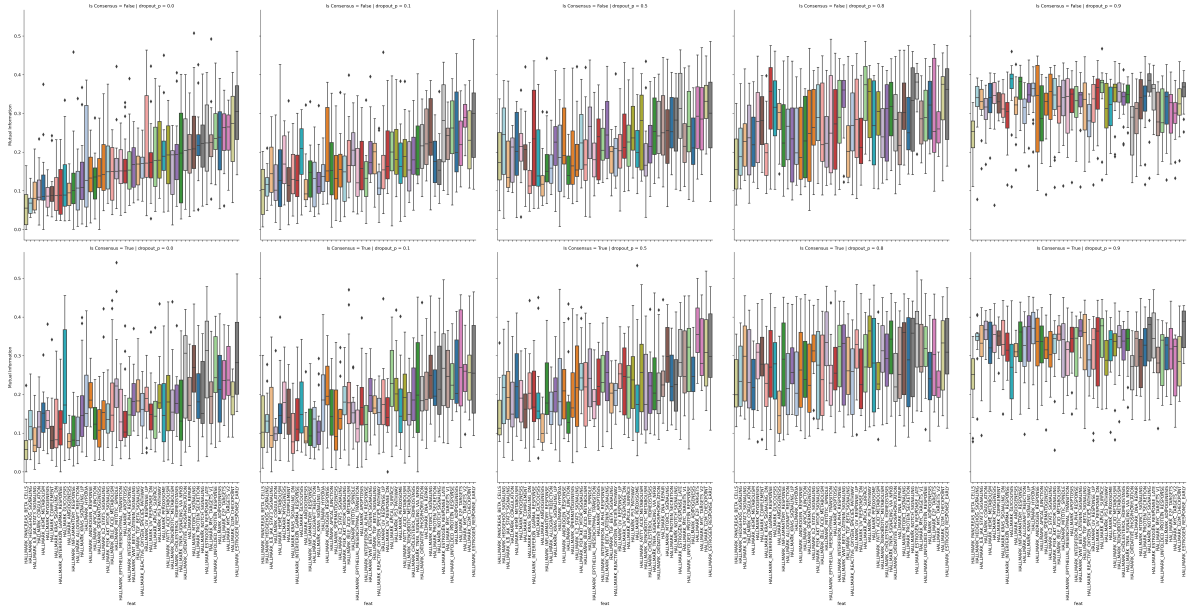}
    \caption{Boxplots of the mutual information between a pathway and PAM50 labels, obtained from models trained with different random seeds on the \texttt{MO-TCGA-BRCA} dataset. Models are labelled as either Aligned Consensus Model (bottom) or not (top). From left to right, we show the results for dropout $p$ being $0.0$, $0.1$, $0.5$, $0.8$, and $0.9$. Note that model order changes drastically at the last two dropout values.}
    \label{fig:dropout-bothmi}
\end{figure}

\begin{figure}
    \centering
    \includegraphics[width=0.95\linewidth]{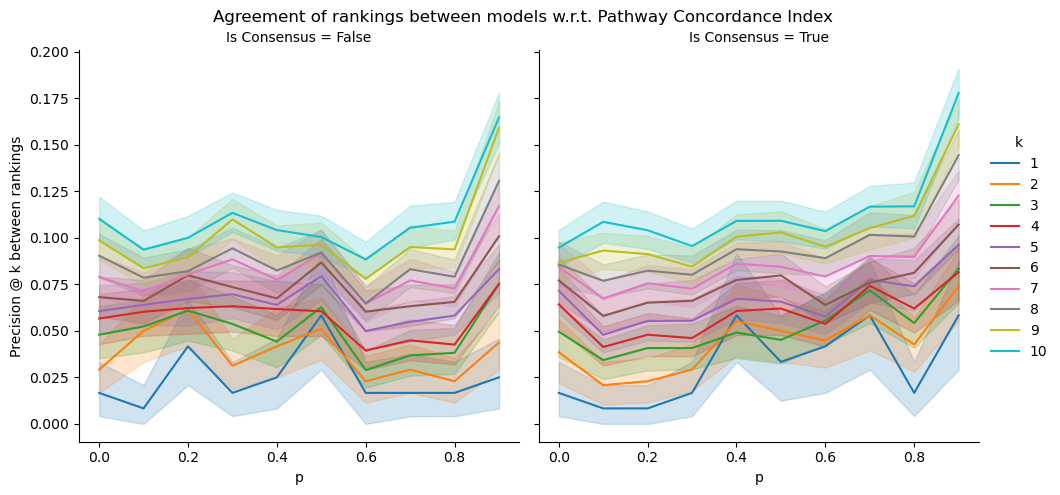}
    \caption{Average top-$k$ agreement of Concordance-Index rankings between models models trained with different random seeds. Models are labelled as either Aligned Consensus Model (right) or not (left). Shaded regions indicate model variance.}
    \label{fig:dropout-rankingci}
\end{figure}

\begin{figure}
    \centering
    \includegraphics[width=0.95\linewidth]{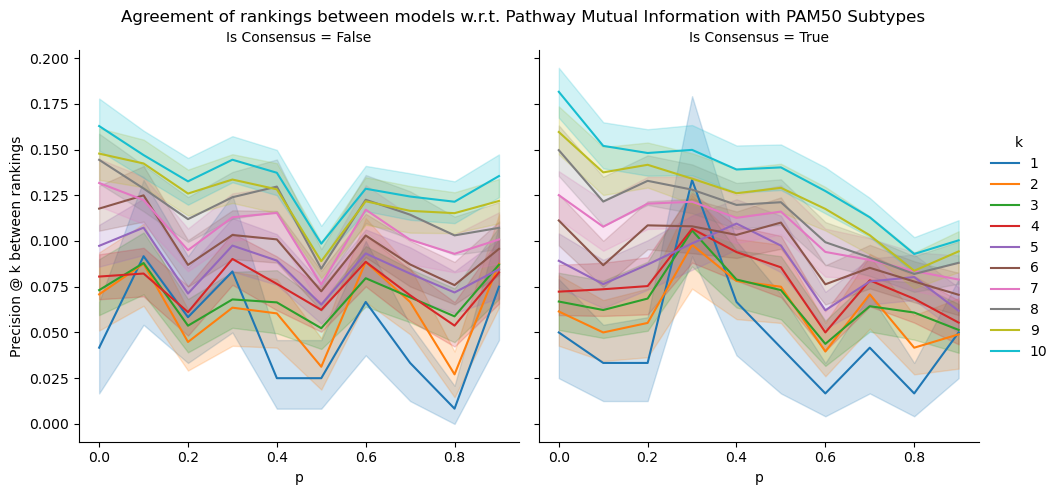}
    \caption{Average top-$k$ agreement the mutual information rankings between models models trained with different random seeds. Models are labelled as either Aligned Consensus Model (right) or not (left). Shaded regions indicate model variance.}
    \label{fig:dropout-rankingmi}
\end{figure}

\begin{figure}
    \centering
    \begin{tabular}{cc}
         \includegraphics[width=0.45\linewidth]{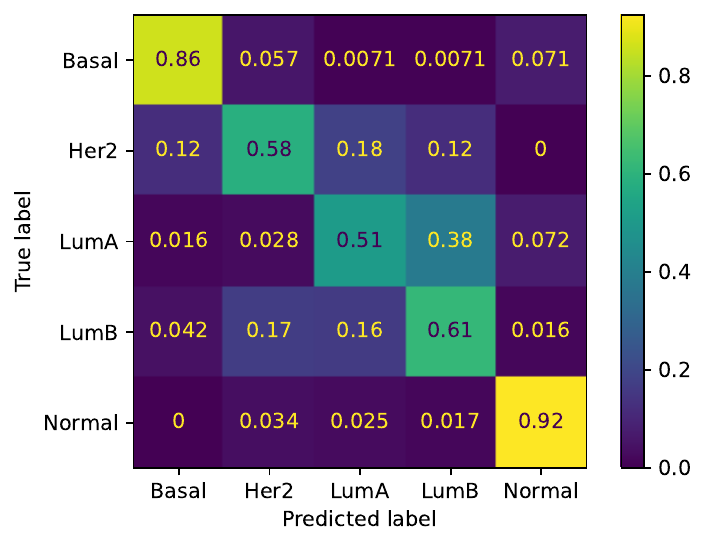} & \includegraphics[width=0.45\linewidth]{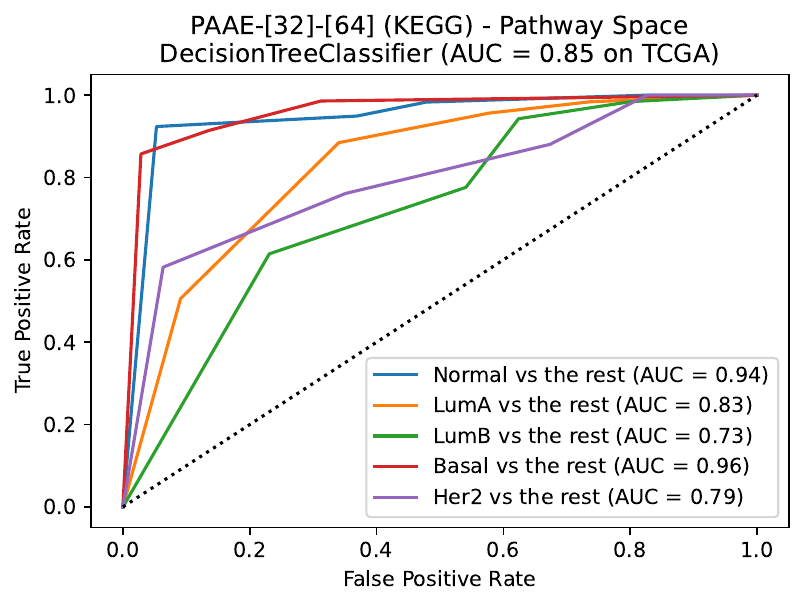} \\
         \includegraphics[width=0.45\linewidth]{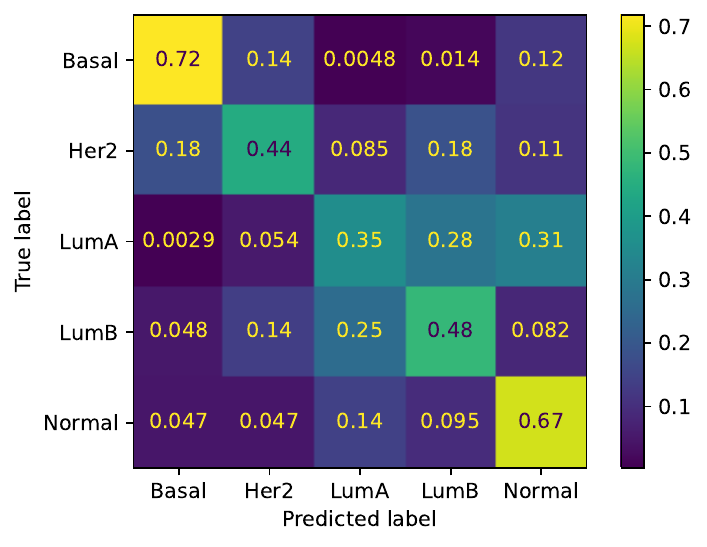} & \includegraphics[width=0.45\linewidth]{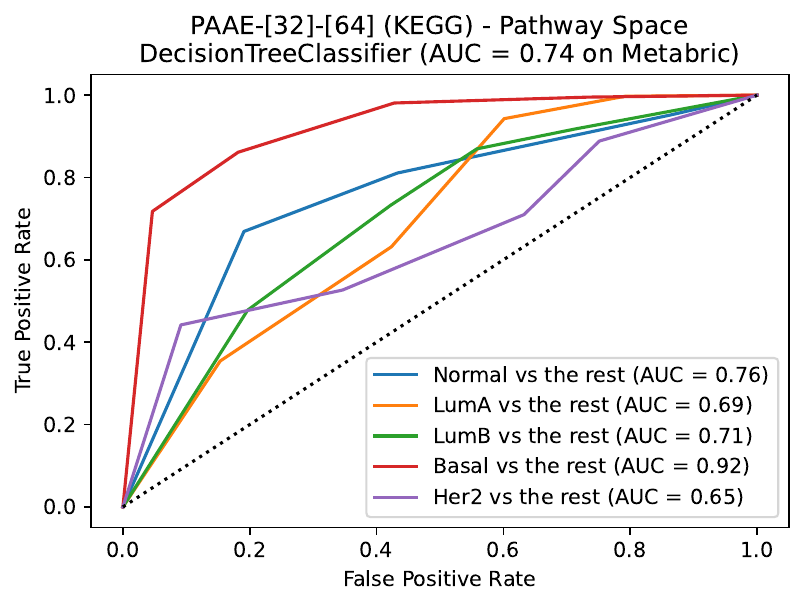} \\
    \end{tabular}
    \caption{Confusion matrices (left) and receiver operating characteristic curves (right) for a constrained decision tree model trained only on 5 KEGG pathways with highest OvR mutual information with each of PAM50 subtypes. Results are  shown for both the TCGA dataset (top, training) and Metabric (bottom, external validation).}
    \label{fig:dtree-validation}
\end{figure}

\begin{figure}
    \centering
    \includegraphics[width=0.95\linewidth]{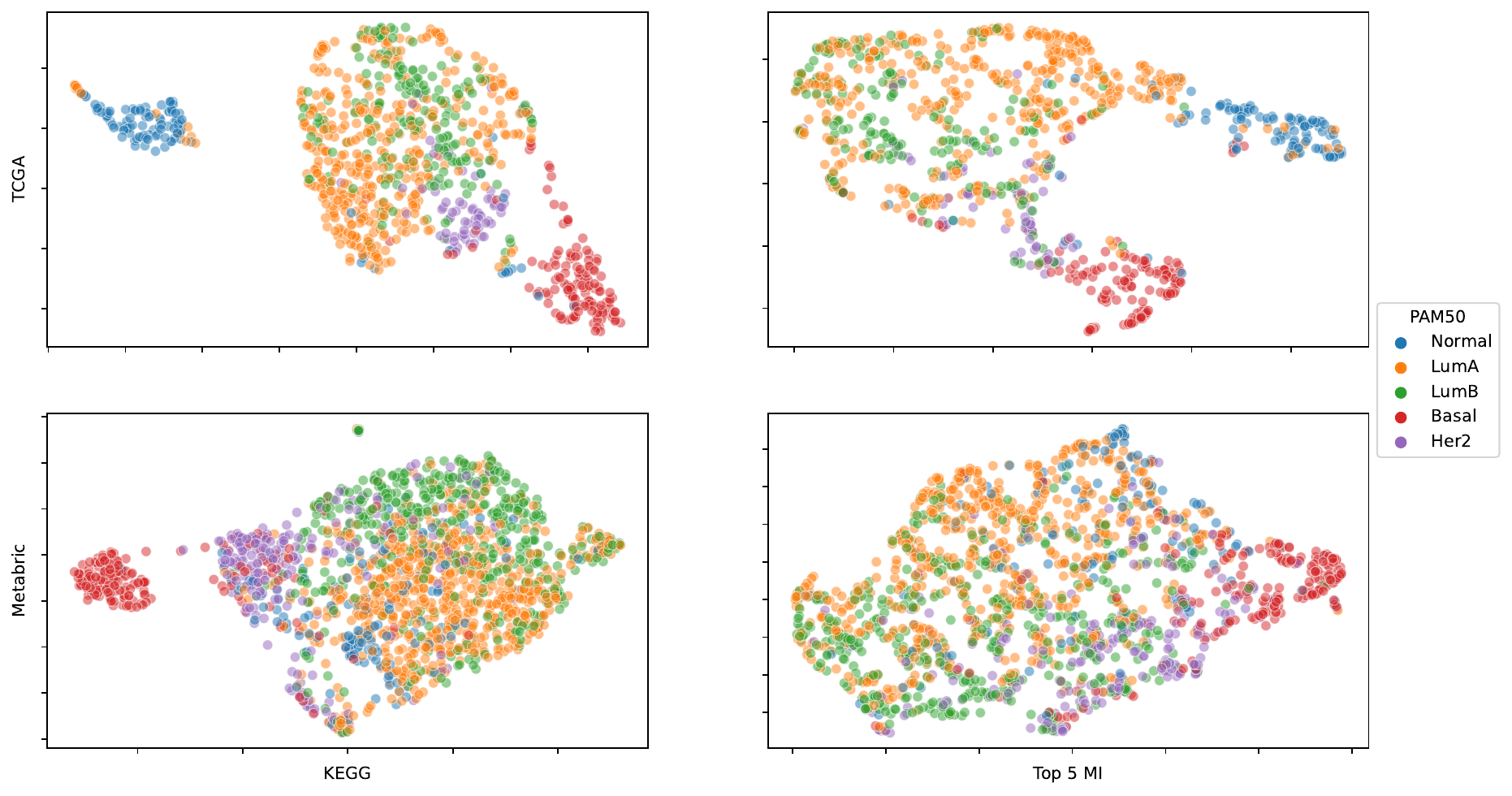}
    \caption{2-dimensional UMAP reprentations of the TCGA (top) and Metabric (bottom) datasets, for the pathway activity vector using all of the KEGG pathways (left) and using the top 5 pathways with highest OvR MI (right).}
    \label{fig:umap-full-vs-ovrmi}
\end{figure}

\begin{figure}
    \centering
    \includegraphics[width=.99\linewidth]{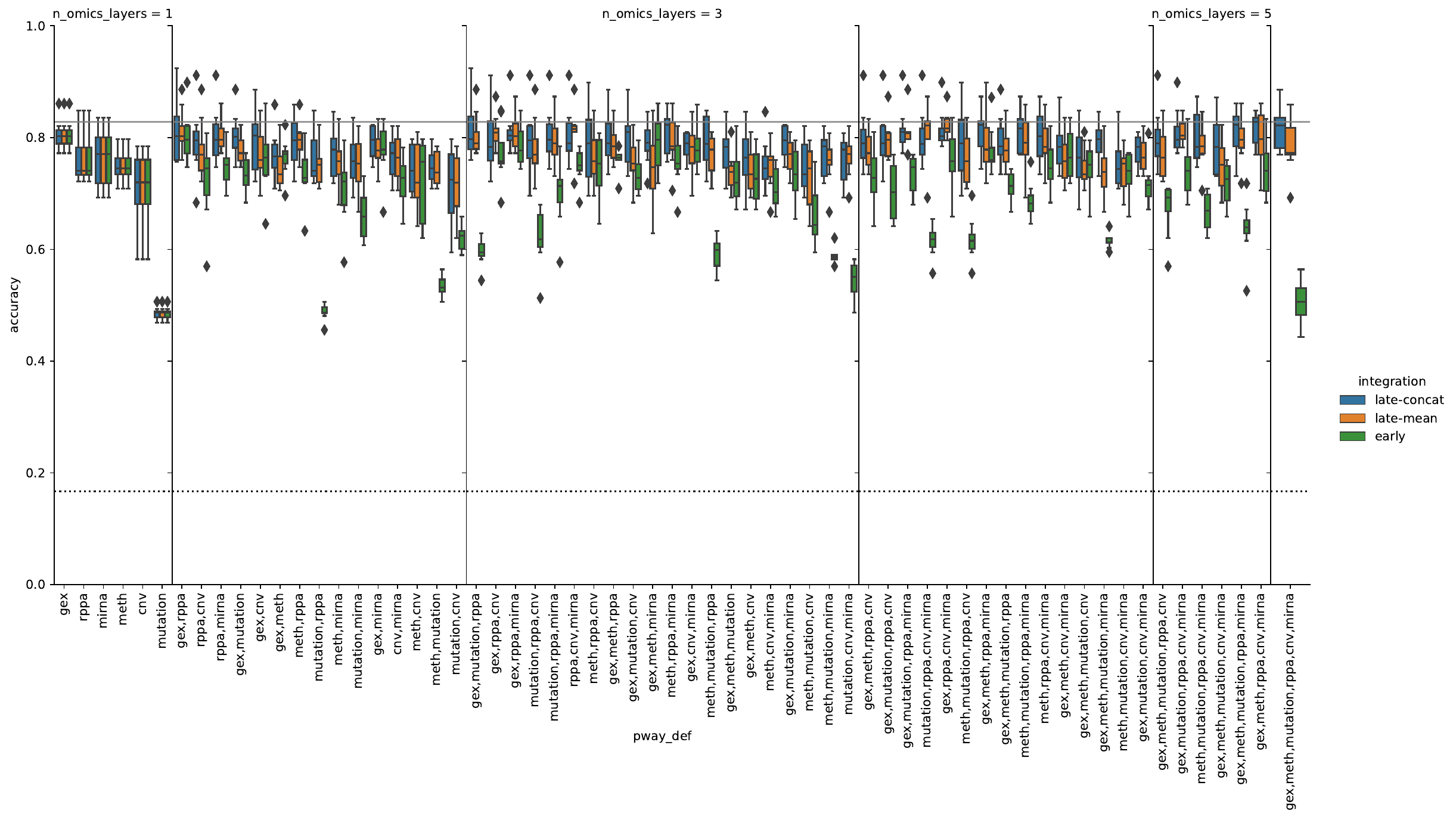}
    \caption{Accuracy of one-vs-rest logistic regression for all combinations of omics layer in early, late-concat and late-mean integration. The gray line indicates the highest median C-index among all tested configurations (dotted line marks accuracy of 0.2, equivalent to random chance).}
    \label{fig:clf-acc}
\end{figure}

\begin{figure}
    \centering
    \includegraphics[width=.99\linewidth]{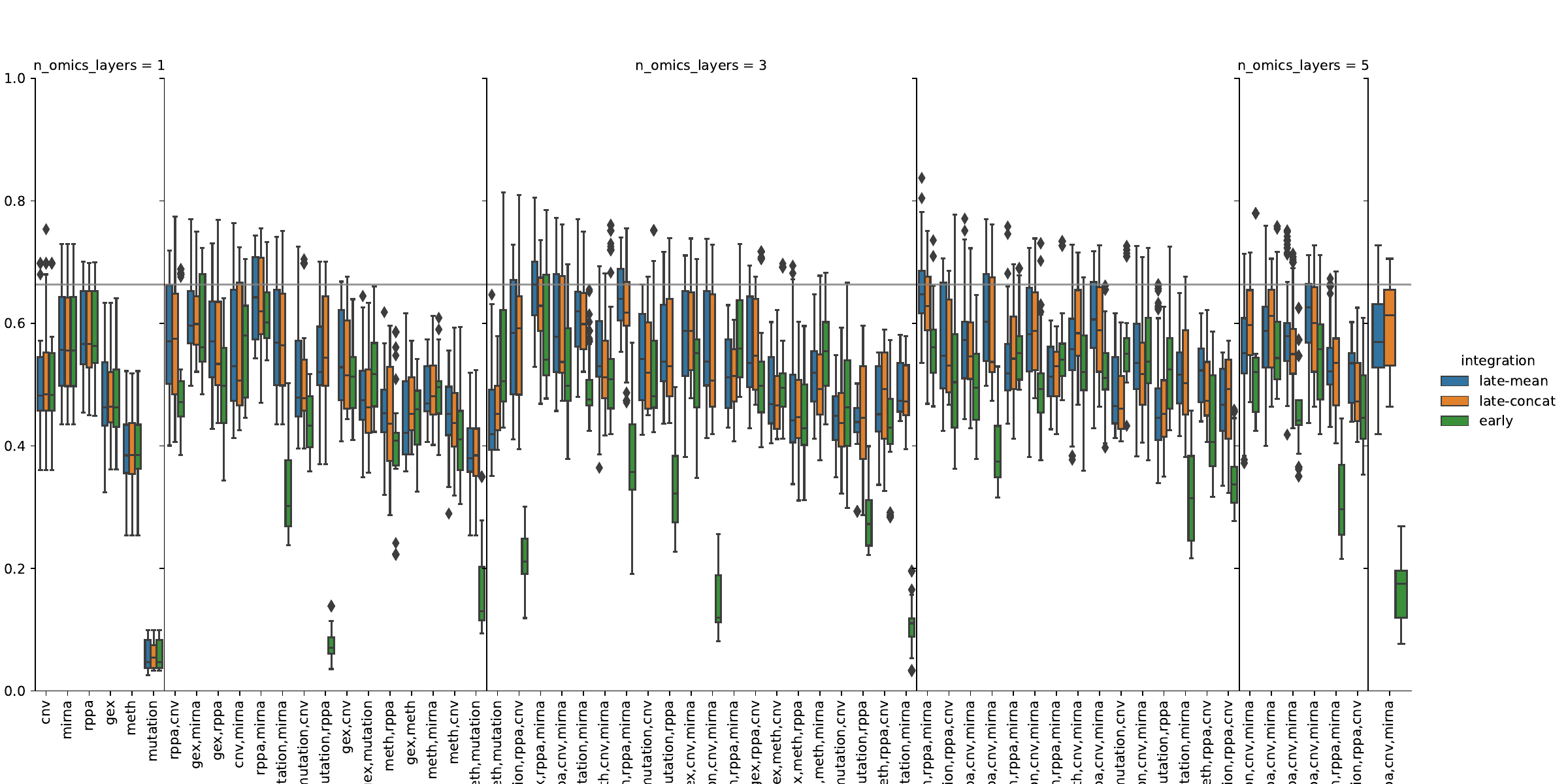}
    \caption{Mutual information between the PAM50 Subtypes and the result of unsupervised clustering using the K-Means algorithm  for all combinations of omics layer in early, late-concat and late-mean integration. The gray line indicates the highest median C-index among all tested configurations.}
    \label{fig:cls-mi}
\end{figure}

\begin{figure}
    \centering
    \includegraphics[width=.99\linewidth]{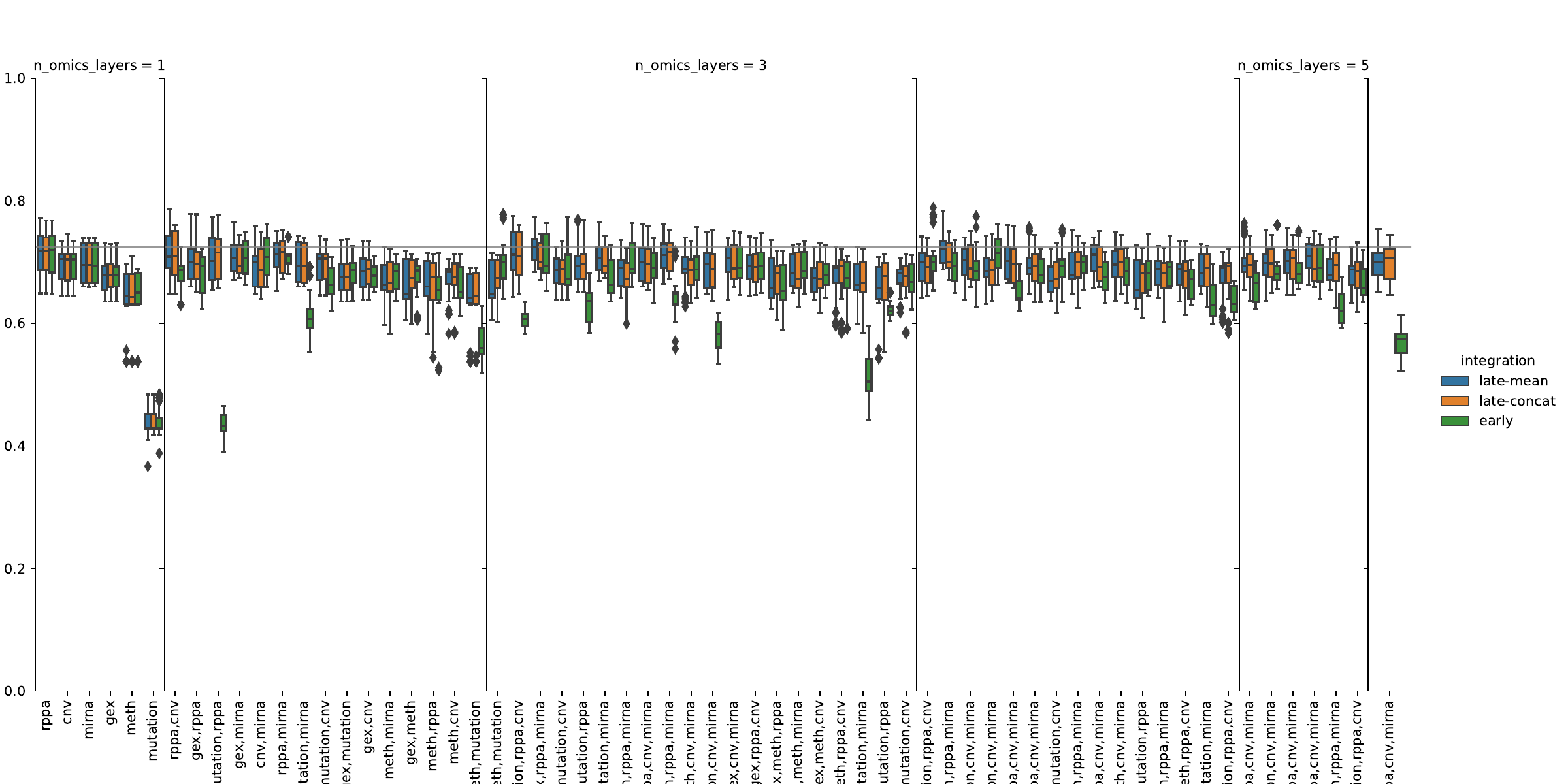}
    \caption{Rand Index between the PAM50 Subtypes and the result of unsupervised clustering using the K-Means algorithm  for all combinations of omics layer in early, late-concat and late-mean integration. The gray line indicates the highest median C-index among all tested configurations.}
    \label{fig:cls-ri}
\end{figure}

\begin{figure}
    \centering
    \includegraphics[width=0.99\linewidth]{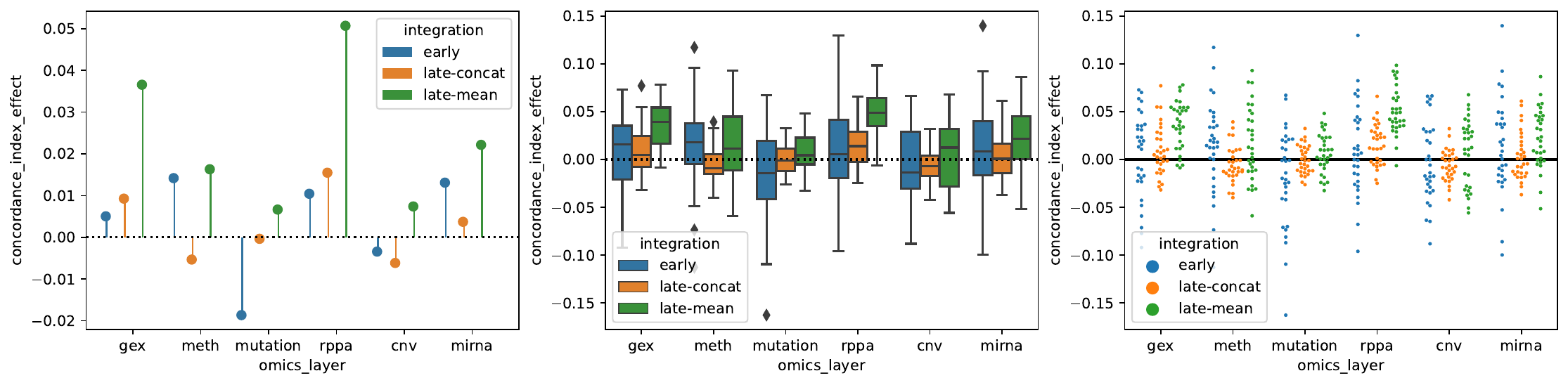}
    \caption{Marginal contribution of each omics layer to survival prediction, measured by concordance index of a Cox proportional hazards model, across the three tested integration methods: early, late-concat and late-mean. The left panel shows the median marginal contribution, the right panel displays a swarm plot of values across all combinations excluding that layer, and in the middle the same information is shown as a boxplot.}
    \label{fig:omics-effect-sur-ci}
\end{figure}

\end{document}